\documentclass[conference]{IEEEtran}
\IEEEoverridecommandlockouts
\usepackage{cite}
\usepackage{amsmath,amssymb,amsfonts}
\usepackage{algorithmic}
\usepackage{graphicx}
\usepackage{textcomp}
\usepackage{xcolor}

\usepackage[hidelinks]{hyperref}
\usepackage{times}
\usepackage{soul}
\usepackage{url}
\usepackage{amsthm}
\usepackage{booktabs}
\usepackage{multirow}
\usepackage{array}
\usepackage{float}
\usepackage{algorithm}
\usepackage{verbatim}
\usepackage{subfigure}
\usepackage{makecell}
\usepackage{balance}
\newtheorem{defn}{Definition}

\newcommand{\Rmnum}[1]{\uppercase\expandafter{\romannumeral #1}}

\def\BibTeX{{\rm B\kern-.05em{\sc i\kern-.025em b}\kern-.08em
    T\kern-.1667em\lower.7ex\hbox{E}\kern-.125emX}}

\begin{document}

\title{Disconnected Emerging Knowledge Graph Oriented Inductive Link Prediction}

\author
{
	Yufeng Zhang{\small$^1$}\hspace*{10pt}
	Weiqing Wang{\small$^2$}\hspace*{10pt}
	Hongzhi Yin{\small$^3$} \hspace*{10pt}
	Pengpeng Zhao{\small$^1$} \hspace*{10pt}
	Wei Chen{\small$^1$}$^*$ \thanks{* These authors are corresponding authors.} \hspace{10pt}
	Lei Zhao{\small$^1$}$^*$\hspace{10pt}\\
	\fontsize{10}{10}\selectfont\itshape $~^1$School of Computer Science and Technology, Soochow University\\
	\fontsize{10}{10}\selectfont\itshape $~^2$ Department of Data Science and AI, Monash University\\
	\fontsize{10}{10}\selectfont\itshape $~^3$ School of Information Technology and Electrical Engineering, The University of Queensland\\
	\fontsize{9}{9}\selectfont\ttfamily\upshape$~^1$yfzhang@stu.suda.edu.cn\hspace*{10pt}$~^2$Teresa.Wang@monash.edu \\
	$~^3$db.hongzhi@gmail.com\hspace*{10pt}$~^1$\{ppzhao,robertchen,zhaol\}@suda.edu.cn
}

\maketitle

\begin{abstract}
Inductive link prediction (ILP) is to predict links for unseen entities in emerging knowledge graphs (KGs), considering the evolving nature of KGs. A more challenging scenario is that emerging KGs consist of only unseen entities without any edge connected to original KGs, called as disconnected emerging KGs (DEKGs). Existing studies for DEKGs only focus on predicting \emph{enclosing links}, i.e., predicting links inside the emerging KG.
The \emph{bridging links}, which carry the evolutionary information from the original KG to DEKG, have not been investigated by previous work so far.
To fill in the gap, we propose a novel model entitled DEKG-ILP (Disconnected Emerging Knowledge Graph Oriented Inductive Link Prediction) that consists of the following two components. (1) The module CLRM (Contrastive Learning-based Relation-specific Feature Modeling) is developed to extract global relation-based semantic features that are shared between original KGs and DEKGs with a novel sampling strategy. (2) The module GSM (GNN-based Subgraph Modeling) is proposed to extract the local subgraph topological information around each link in KGs. The extensive experiments conducted on several benchmark datasets demonstrate that DEKG-ILP has obvious performance improvements compared with state-of-the-art methods for both enclosing and bridging link prediction.
\end{abstract}


\section{Introduction}
\label{sec:1}

Knowledge Graphs (KGs), such as Freebase\cite{Freebase}, NELL\cite{NELL}, and DBpedia\cite{DBpedia} play a critical role in many applications like information retrieval \cite{KG_IR_1,KG_IR_2}, recommendation systems \cite{add2,add4}, multi-hop query \cite{add5}, and question answering\cite{KG_QA_1,KG_QA_2}.
A typical KG models data as a collection of facts and specify entities as nodes, relations as edges, having a strong ability to represent structured data.
Predicting missing facts in KGs, also known as KG link prediction, is a widely studied problem \cite{KG_review} and has been proven successful benefiting from recent KG embedding methods \cite{TransE,ConvE,R-GCN}.

\begin{figure}
	\centering
	\includegraphics[width=\linewidth]{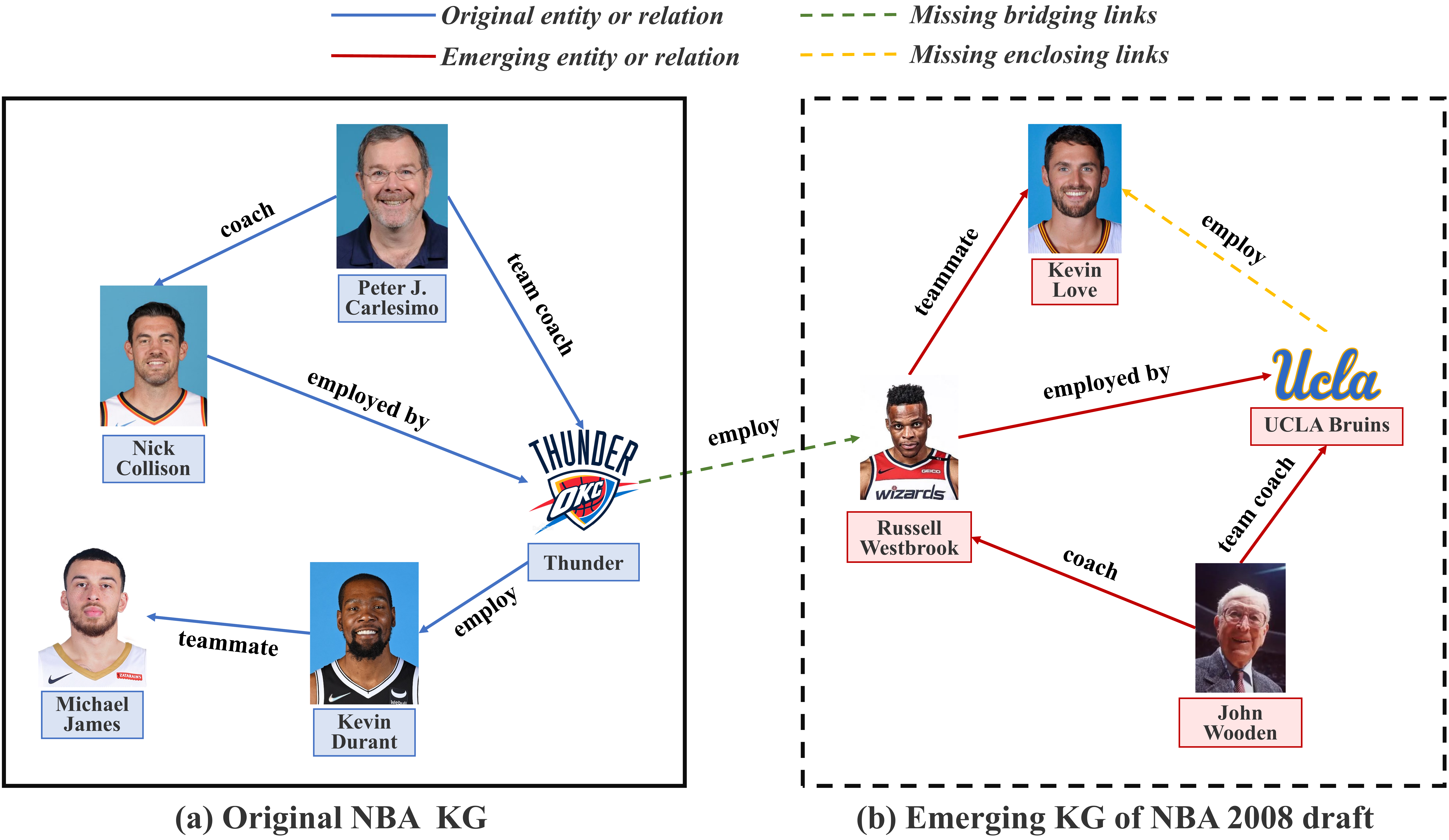}
	\vspace{-15pt}
	\caption{A motivating example of inductive link prediction. The original KG in left and DEKG in right are two disconnected KGs without any edges connecting them.}
	\vspace{-15pt}
	\label{fig:Relevance_of_KG}
\end{figure}

Despite the success, KG link prediction remains challenging in many real-world scenarios. Shi and Weninger reported \cite{ConMask} that KGs are dynamically evolving rather than staying static, e.g., around 200 unseen entities emerged every day during late 2015 and early 2016 in DBpedia.
However, the traditional transductive KG embedding methods are ineffective for emerging KGs as the new entities are unseen during training. Although this problem can be solved by retraining models with the whole graph combined with unseen elements, the consumption of time and computation is intolerable in real-world applications \cite{KG_review,MEAN}. To address this problem, increasing attention has been paid to inductive link prediction (ILP), which aims to predict links for unseen entities in the emerging KGs. 
Several graph neural network based methods \cite{MEAN,LAN,GEN} have been proposed by transferring information from original KGs to emerging KGs to obtain the embeddings of unseen entities without retraining the whole KG.

Recently, \cite{Grail} introduced a more challenging scenario where emerging KGs consist of unseen entities only and the edges between the original KGs and emerging KGs are not observed, called as disconnected emerging KGs (DEKGs) scenario. To be noted, the relation space is shared between the original KGs and DEKGs (i.e., there are only unseen entities and no unseen relations in DEKGs). 
For this scenario, Grail\cite{Grail} and TACT\cite{TACT} have been proposed to predict the \emph{enclosing links}, where both the head and tail entities are unseen entities inside DEKGs, by learning logical rules and reasoning over subgraph structures in an entity-independent manner. Despite the great contributions made by \cite{Grail} and \cite{TACT}, the prediction of links connecting  DEKGs and original KGs, which are formulated as \emph{bridging links}, has not been exploited so far.

Fig.~\ref{fig:Relevance_of_KG} presents a motivating example of bridging link prediction for DEKGs. It describes KGs of the NBA 2008 draft, where the DEKG in Fig.~\ref{fig:Relevance_of_KG}(b) is composed of new entities and shares the common relation space with the original KG in Fig.~\ref{fig:Relevance_of_KG}(a). 
In this example, the participation of \emph{Russell} to \emph{Thunder}, i.e., the bridging link (\emph{Thunder}, \emph{employ}, \emph{Russel}), brought a significant benefit to \emph{Thunder} in the following seasons.
Generally, the absence of \emph{bridging links} between two disconnected KGs is common in real-world applications, while these links usually imply some critical information, such as the drug-drug interaction that helps develop new medicine (e.g., the discovery of Artemisinin) or the connection between two cases that was ignored by the police. 
Actually, the motivation of this paper also comes from a real-world criminal case in 2015\footnote{\url{https://baijiahao.baidu.com/s?id=1718118167319289628}}. A neglected connection between the case and another seemingly unrelated one that happened several years ago brought a significant breakthrough in cracking both cases.
In summary, revealing the connection between original KGs and DEKGs (i.e., predicting \emph{bridging links}) can benefit many cross-graph applications \cite{KG_review,add1,add3}.

However, existing studies for DEKGs (i.e., Grail \cite{Grail} and TACT \cite{TACT}) cannot handle this problem effectively. This is because their target is to predict \emph{enclosing links} (i.e., yellow dashed link in Fig.~\ref{fig:Relevance_of_KG}) inside DEKGs using the subgraph reasoning method, which seriously suffers from the \emph{topological limitation} in DEKGs. 
Specifically, the \emph{topological limitation} means there is no \textbf{connected} subgraph around a bridging link.
Taking the target bridging link (\emph{Thunder}, \emph{employ}, \emph{Russell}) in Fig.~\ref{fig:Relevance_of_KG} as an example, the head entity \emph{Thunder} and the tail entity \emph{Russell} are in two disconnected KGs respectively. So the subgraphs constructed for this target link are two disconnected subgraphs where one is around \emph{Thunder} in the DEKG and the other one is around \emph{Russell} in the original KG, and there is no path connecting the two subgraphs. 
Unfortunately, the underlying idea of Grail \cite{Grail} and TACT \cite{TACT} relies on the connectivity between two entities to perform path reasoning over a connected subgraph. For example, they rely on the relational path (\emph{Kevin~Love} $\rightarrow$ \emph{Russell} $\rightarrow$ \emph{UCLA~Bruins}) to predict the link (\emph{UCLA~Bruins}, \emph{employ}, \emph{Kevin~Love}), while the path does not exist for a bridging link as we discussed above. Consequently, both Grail \cite{Grail} and TACT \cite{TACT} are problematic in predicting \emph{bridging links} for DEKGs.

To deal with the above-mentioned problem, we propose a novel model namely DEKG-ILP, which contains two modules: Contrastive Learning-based Relation-specific Feature Modeling (CLRM) and GNN-based Subgraph Modeling (GSM). 
Both CLRM and GSM are carefully designed to deal with the \emph{topological limitation} problem, where CLRM is a fundamentally novel module to exploit global semantic information in KGs, and GSM is an improved method developed from Grail \cite{Grail} to extract local topological around links.
Specifically, the module CLRM first extracts global relation-based semantic information shared between original KGs and DEKGs and represents entities in an entity-independent manner. 
The key idea is that the semantic representation of an entity is computed based on its associated relations (e.g., \emph{Russell} is recognized as an \emph{Employee} and a \emph{Sports player} since he is associated with the relations \emph{teammate}, \emph{employed by}, and \emph{coach}).
Following this intuition, the feature for each relation is defined and then entities can be represented as a fusion of the relation corresponding features. 
In this way, the entities in original KGs and DEKGs are linked via the shared relation space rather than topological graph structure, thus tackling the \emph{topological limitation}.
Additionally, a novel contrastive learning enabled sampling strategy is designed to generate positive and negative examples for each entity to optimize the relation-specific features.
Secondly, the GNN-based subgraph modeling module GSM is employed to exploit the local topological information of the subgraph around each link in KGs. A novel node labeling method is proposed in GSM to simulate the disconnected nodes and deal with the \emph{topological limitation} problem. Compared with existing studies \cite{Grail,TACT} that only focus on the prediction of \emph{enclosing links}, both \emph{bridging links} and \emph{enclosing links} are exploited in this work.

To summarize, our main contributions are as follow:
\begin{itemize}
\item We extend the existing formulation of inductive link prediction for unseen entities in a disconnected emerging scenario, by considering \emph{enclosing links} and especially \emph{bridging links} simultaneously.
\item We propose a novel model DEKG-ILP which can effectively solve the extended inductive link prediction task. Two carefully designed modules are included in DEKG-ILP. Firstly, the module CLRM is used to extract global relation-based semantic features that are shared between original KGs and DEKGs, where a contrastive learning method is employed to optimize these features. Secondly, a GNN-based subgraph modeling module GSM is used to exploit the local topological information around each link in KGs.
\item The comprehensive experiments conducted on several benchmark datasets demonstrate that our proposed model DEKG-ILP outperforms existing methods on predicting \emph{enclosing links}. Moreover, different from existing methods, DEKG-ILP is able to predict \emph{bridging links}.
\end{itemize}

The rest of the paper is organized as follows. Section~\ref{sec:related_work} reviews the related work of KG link prediction and contrastive learning methods. Section~\ref{sec:3} introduces the basic concepts and formulates the problem. Section~\ref{sec:method} firstly provides an overview of our proposed model DEKG-ILP and then describes the model in detail. We provide our experimental setup and results in Section~\ref{sec:experiments} and  conclude this work in Section~\ref{sec:conclusion}.

\begin{table*}[]
\label{tab:related_work}
\centering
\caption{Summary of KG link prediction methods. \checkmark means being able to handle this task and $\times$ means not.}
    \begin{tabular}{cccccc}
    \hline
    \multicolumn{1}{c|}{\multirow{3}{*}{}} & \multicolumn{1}{c|}{\multirow{3}{*}{\textbf{Model}}} & \multicolumn{1}{c|}{\multirow{3}{*}{\textbf{\makecell[c]{Transductive\\Link Prediction}}}} & \multicolumn{3}{c}{\textbf{Inductive Link Prediction}} \\ 
    \cline{4-6} 
    \multicolumn{1}{c|}{} & \multicolumn{1}{c|}{} & \multicolumn{1}{c|}{} & \multicolumn{1}{c|}{\multirow{2}{*}{\makecell[c]{Common\\Emerging KG}}} & \multicolumn{2}{c}{Disconnected Emerging KG} \\ 
    \cline{5-6} 
    \multicolumn{1}{c|}{} & \multicolumn{1}{c|}{} & \multicolumn{1}{c|}{} & \multicolumn{1}{c|}{} & \multicolumn{1}{c|}{Enclosing Link} & \multicolumn{1}{c}{Bridging Link} \\ 
    \hline
    \multicolumn{1}{c|}{\multirow{3}{*}{Transductive Methods}} & \multicolumn{1}{c|}{TransE \cite{TransE}} & \multicolumn{1}{c|}{\checkmark} & \multicolumn{1}{c|}{$\times$} & \multicolumn{1}{c|}{$\times$} & \multicolumn{1}{c}{$\times$} \\
    \cline{2-6} 
    \multicolumn{1}{c|}{} & \multicolumn{1}{c|}{RotatE \cite{RotatE}} & \multicolumn{1}{c|}{\checkmark} & \multicolumn{1}{c|}{$\times$} & \multicolumn{1}{c|}{$\times$} & \multicolumn{1}{c}{$\times$} \\ 
    \cline{2-6} 
    \multicolumn{1}{c|}{} & \multicolumn{1}{c|}{ConvE \cite{ConvE}} & \multicolumn{1}{c|}{\checkmark} & \multicolumn{1}{c|}{$\times$} & \multicolumn{1}{c|}{$\times$} & \multicolumn{1}{c}{$\times$} \\
    \hline
    \multicolumn{1}{c|}{\multirow{7}{*}{Inductive Methods}} & \multicolumn{1}{c|}{MEAN \cite{MEAN}} & \multicolumn{1}{c|}{\checkmark} & \multicolumn{1}{c|}{\checkmark} & \multicolumn{1}{c|}{$\times$} & \multicolumn{1}{c}{$\times$} \\
    \cline{2-6} 
    \multicolumn{1}{c|}{} & \multicolumn{1}{c|}{GEN \cite{GEN}} & \multicolumn{1}{c|}{\checkmark} & \multicolumn{1}{c|}{\checkmark} & \multicolumn{1}{c|}{$\times$} & \multicolumn{1}{c}{$\times$} \\ 
    \cline{2-6} 
    \multicolumn{1}{c|}{} & \multicolumn{1}{c|}{Neural LP \cite{NeuralLP}} & \multicolumn{1}{c|}{\checkmark} & \multicolumn{1}{c|}{\checkmark} & \multicolumn{1}{c|}{\checkmark} & \multicolumn{1}{c}{$\times$} \\ 
    \cline{2-6} 
    \multicolumn{1}{c|}{} & \multicolumn{1}{c|}{RuleN \cite{RuleN}} & \multicolumn{1}{c|}{\checkmark} & \multicolumn{1}{c|}{\checkmark} & \multicolumn{1}{c|}{\checkmark} & \multicolumn{1}{c}{$\times$} \\ 
    \cline{2-6} 
    \multicolumn{1}{c|}{} & \multicolumn{1}{c|}{Grail \cite{Grail}} & \multicolumn{1}{c|}{\checkmark} & \multicolumn{1}{c|}{\checkmark} & \multicolumn{1}{c|}{\checkmark} & \multicolumn{1}{c}{$\times$} \\ 
    \cline{2-6} 
    \multicolumn{1}{c|}{} & \multicolumn{1}{c|}{TACT \cite{TACT}} & \multicolumn{1}{c|}{\checkmark} & \multicolumn{1}{c|}{\checkmark} & \multicolumn{1}{c|}{\checkmark} & \multicolumn{1}{c}{$\times$} \\ 
    \cline{2-6} 
    \multicolumn{1}{c|}{} & \multicolumn{1}{c|}{DEKG-ILP} & \multicolumn{1}{c|}{\checkmark} & \multicolumn{1}{c|}{\checkmark} & \multicolumn{1}{c|}{\checkmark} & \multicolumn{1}{c}{\checkmark} \\ 
    \hline
    \end{tabular}
\end{table*}

\section{Related Work}
The related studies, which contain the methods of link prediction and contrastive learning, are presented in this section. Additionally, TABLE~\ref{tab:related_work} provides a summary of what tasks these link prediction methods can handle respectively.
\label{sec:related_work}

\subsection{Transductive Link Prediction Methods.}
Transductive embedding methods require that all entities can be obtained during training. Translational distance based methods \cite{TransE,TransH,TransN} measure the plausibility of facts as the distance between entities carried out by relations. \cite{RotatE,HAKE} further improve the translational distance based methods by modeling the distance translation between entities as a rotation in the complex space. Factorization based models \cite{RESCAL,DistMult,ComplEx} extract the latent semantics of entities and relations in their embedding space. Recently, \cite{ConvE,ConvKB} reshape KG embedding and employ CNN as the encoder. Some studies \cite{R-GCN,SACN} also employ GNNs to aggregate neighborhood information to exploit structural information in the graph. Unfortunately, all the above methods have to retrain the whole graph when new KGs emerge and cannot generalize to unseen entities. However, the time and computational cost of retraining is intolerable in real-world applications \cite{KG_review,MEAN}.

\subsection{Inductive Link Prediction Methods.}
\label{sec:inductive methods}

\subsubsection{Additional Information Methods}
Inductive methods aim to predict links for unseen entities in emerging KGs without retraining the whole graph. 
Some works \cite{TKRL,IKRL} embed unseen entities using additional information like text descriptions or images. However, these methods could be limited when the additional information is missing or insufficient, which is usually the case in real-world KGs. 
Moreover, constructing additional information for KGs also costs. Thus the additional information based method is not a good solution to ILP.

\subsubsection{Rule Induction Methods}
Rule learning based methods \cite{AMIE,RuleN,RLvRL} induce probabilistic logical rules by enumerating statistical regularities and patterns present in the knowledge graph and are inherently inductive since the rules are independent of node identities. The traditional rule-based method \cite{traditional_rule} mines rules from data by inductive logic programming but suffers from the problem of scaling to large datasets and being challenging to optimize. Recently, Neural LP \cite{NeuralLP} proposes an end-to-end differentiable framework to learn rules using TensorLog \cite{TensorLog} operators. DRUM \cite{DURM} further improves Neural LP by mining more accurate logical rules. However, rule learning based methods mainly focus on mining horn rules, limiting their ability to model more complex semantic correlations between relations in knowledge graphs.

\subsubsection{Embedding based Methods}
Several GNN-based methods \cite{MEAN,LAN,GEN} obtain the embeddings of unseen entities by aggregating information from original KGs to emerging KGs.
MEAN \cite{MEAN} employs GNNs to encode unseen entities by aggregating information from its neighbors in original KGs with a simple pooling function. LAN \cite{LAN} introduces two attention mechanisms for the GNN in their model, where the attention weights are computed with logic rules and learned using neural networks respectively. VN network \cite{VNnetwork} is proposed to solve the data sparsity problem by constructing virtual neighbors for unseen entities.
GEN \cite{GEN} further employs a meta-learning framework to simulate the emerging KGs scenario during training to make the model able to generalize to unseen entities inherently.
Recently, Grail \cite{Grail} introduces a more challenging scenario that we termed as DEKGs. In this scenario, all the above-mentioned methods are problematic as they are based on the graph message passing methods, which depend on the edges that exist between the original and emerging KGs, thus conflict with the scenario of DEKGs. 
To handle this problem, Grail \cite{Grail} and TACT \cite{TACT} are proposed to investigate the problem of enclosing link prediction in DEKGs by reasoning over local subgraph structures. Despite the great contributions made by Grail and TACT, their proposed methods are not suitable for predicting \emph{bridging links}.

\subsection{Contrastive Learning in Graph.}
Contrastive learning is a class of self-supervised methods and has been applied in many CV and NLP applications\cite{word2vector,MoCo,contrastive_NLP1}. Recently, contrastive learning can be found in several graph representation learning algorithms. DGI \cite{DGI} extends deep Infomax \cite{DeepInfomax} via contrasting node and graph encoding. 
\cite{GCC,DBLP:conf/aaai/ZengX21,Cuco,MoCL} employ contrastive learning methods and generate positive and negative examples based on topological information of different graph structures. However, KGs contain not only topological information, thus we propose a novel sampling strategy for contrastive learning to fully utilize the semantic information in KGs, aiming to obtain better representations of entities.

Overall, KG link prediction in both transductive and inductive scenarios has been studied by previous work from many different perspectives, yet no study considers the \emph{bridging links} in the DEKG scenario. In addition, there is no contrastive learning method specially designed for KGs to fully exploit the semantic information in KGs. Consequently, we propose the model DEKG-ILP in this work to consider both \emph{enclosing links} and \emph{bridging links} with a novel contrastive learning sample strategy specially designed for KGs. 

\section{Problem Definition}
\label{sec:3}
In this section, we present definitions used throughout the paper and formulate the extended inductive link prediction problem in a disconnected emerging scenario.

\begin{defn}[Knowledge Graph]
Let $\mathcal{E}$ denotes the set of entities and $\mathcal{R}$ denotes the set of relations. A knowledge graph $\mathcal{G}(\mathcal{E}, \mathcal{R})$ (e.g., Fig.~\ref{fig:Relevance_of_KG}(a)) models data as a collection of triplets $(h,r,t)$, where $h, t \in \mathcal{E}$ and $r \in \mathcal{R}$. Accordingly, a KG can be denoted as $\mathcal{G}(\mathcal{E}, \mathcal{R}) = \{(h, r, t) | h, t \in \mathcal{E}, r \in \mathcal{R}\} \subseteq \mathcal{E} \times \mathcal{R} \times \mathcal{E}$.
\end{defn}

\begin{defn}[Disconnected Emerging Knowledge Graph]
In a disconnected emerging scenario, an emerging KG $\mathcal{G}'(\mathcal{E}', \mathcal{R})$ (e.g., Fig.~\ref{fig:Relevance_of_KG}(b)) consists of unseen entity set $\mathcal{E}'$ and common relation set $\mathcal{R}$ shared with the original KG $\mathcal{G}(\mathcal{E}, \mathcal{R})$, without edges being observed between the two KGs. Formally, a DEKG can be denoted as $\mathcal{G}'(\mathcal{E}', \mathcal{R}) = \{(h, r, t) | h, t \in \mathcal{E}', r \in \mathcal{R}\} \subseteq \mathcal{E}' \times \mathcal{R} \times \mathcal{E}'$, where $\mathcal{E} \cap \mathcal{E}' = \emptyset$.
\end{defn}

\begin{defn}[Enclosing Links]
The \emph{enclosing links} (e.g., the yellow dashed link in Fig.~\ref{fig:Relevance_of_KG}) represent the links between unseen entities in the emerging KG $\mathcal{G}'(\mathcal{E}', \mathcal{R})$, where both the head entity $h$ and tail entity $t$ are unseen. Formally, an enclosing link is denoted as $(h, r, t) \in \mathcal{E}' \times \mathcal{R} \times \mathcal{E}'$.
\end{defn}

\begin{defn}[Bridging Links] 
The \emph{bridging links} (e.g., the green dashed link in Fig.~\ref{fig:Relevance_of_KG}) refer to the links that bridge $\mathcal{G}(\mathcal{E}, \mathcal{R})$ and $\mathcal{G}'(\mathcal{E}', \mathcal{R})$, where one of the head or tail entity is known and the other is unknown. Formally, a bridging link is denoted as $(h, r, t) \in \mathcal{E}' \times \mathcal{R} \times \mathcal{E} \cup \mathcal{E} \times \mathcal{R} \times \mathcal{E}'$.
\end{defn}

\begin{defn}[Problem Formulation]
Given an original KG $\mathcal{G}(\mathcal{E}, \mathcal{R})$ and a DEKG $\mathcal{G}'(\mathcal{E}', \mathcal{R})$, the extended inductive link prediction task aims to predict both bridging links and enclosing links for unseen entities in $\mathcal{G}'(\mathcal{E}', \mathcal{R})$. Specifically, we perform link prediction for each link $(h, r, t)$ in all the forms of $(?, r, t)$, $(h, ?, t)$, $(h, r, ?)$, where the link $(h, r, t)$ can be either an enclosing link or a bridging link.
\end{defn}

\begin{figure*}[ht]
	\centering
	\includegraphics[width=\linewidth]{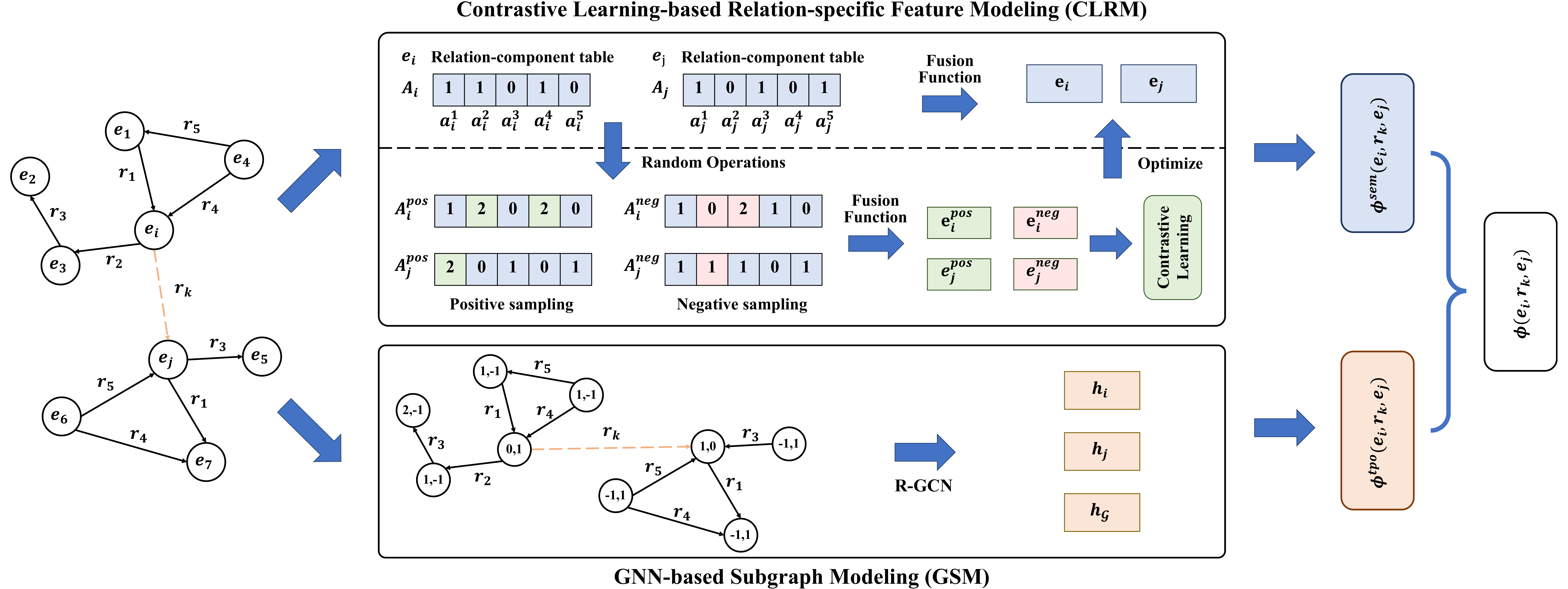}
	\vspace{-10pt}
	\caption{Overview of DEKG-ILP. CLRM shows the construction of relation-component tables for the head and tail entities of the target link, as well as the sampling process for the two entities in contrastive learning. GSM outputs $\boldsymbol{h}_i$, $\boldsymbol{h}_j$, $\boldsymbol{h}_\mathcal{G}$ as the topological representation for $e_i$, $e_j$, and the whole graph.}
	\label{fig:Model}
	\vspace{-10pt}
\end{figure*}

\section{Proposed Model DEKG-ILP}
\label{sec:method}
In this section, we firstly introduce the model overview of the proposed model DEKG-ILP, which consists of two different modules CLRM and GSM. Then the two modules and model training object are discussed in detail.

\subsection{Model Architecture Overview}
The overview of our proposed model DEKG-ILP, which consists of two modules CLRM and GSM, is presented in Fig.~\ref{fig:Model}. Specifically, CLRM extracts global relation-based semantic features shared across KGs, where the contrastive learning is employed to optimize the extracted features with a carefully designed sampling strategy. GSM exploits the local topological information around each link in KGs. 
To predict a target link (e.g., the orange dashed link in Fig.~\ref{fig:Model}), CLRM embeds the head and tail entities (i.e., $\bold{e}_i$ and $\bold{e}_j$) with only the entities' directly associated relations and calculates the score $\phi^{sem} (e_i, r_k, e_j)$ from the semantic perspective. GSM considers the multi-hop subgraph around the target link to embed head entity, tail entity, as well as the whole subgraph (i.e., $\boldsymbol{h}_i$, $\boldsymbol{h}_j$, $\boldsymbol{h}_{\mathcal{G}}$) and calculates the score $\phi^{tpo} (e_i, r_k, e_j)$ from the topological perspective. The final score $\phi(e_i, r_k, e_j)$ for the target link is output by combining the scores from two different modules.

\subsection{Module CLRM} 
\begin{figure}
	\centering
	\includegraphics[width=\linewidth]{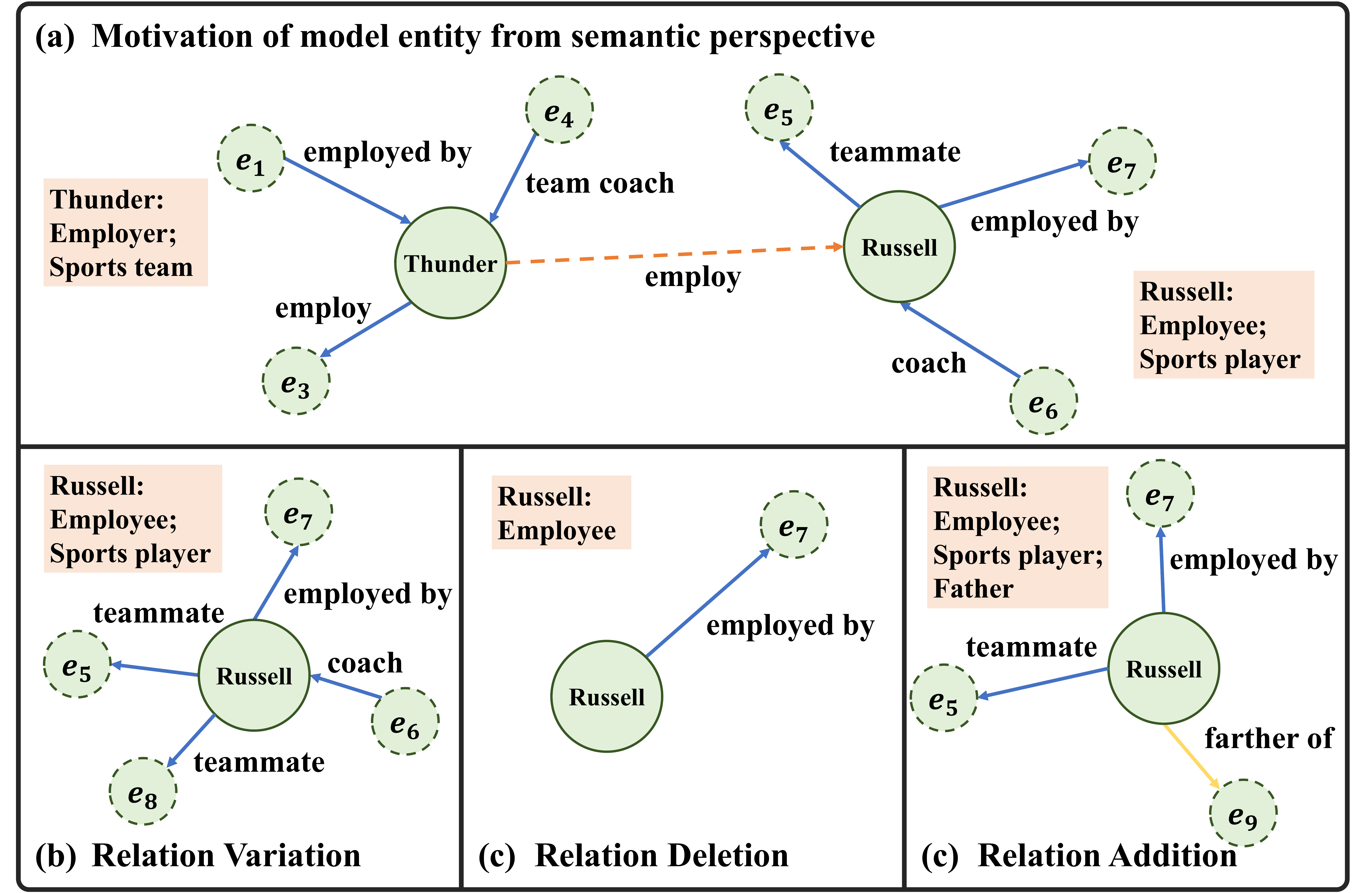}
	\vspace{-10pt}
	\caption{A example of CLRM to illustrate how this module represents entities in an entity independent manner and how the semantic-aware sample strategy generates positive and negative examples for each entity.}
	\label{fig:CLRM_example}
	\vspace{-10pt}
\end{figure}

In this module, the Relation-specific Feature Modeling is designed to extract semantic features for relations shared by original KGs and DEKGs from a global perspective, and represent entities with these features in an entity-independent manner. Furthermore, a contrastive learning based method is employed to optimize these features with a novel semantic-aware sampling strategy to fully exploit the semantic information in KGs. An example of the motivation behind CLRM is given in Fig.~\ref{fig:CLRM_example}.

\subsubsection{Relation-specific Feature Modeling}
In KGs, intuitively, the semantic component of an entity, i.e., what the entity consists of from a semantic perspective, is influenced by its associated relations.
Continuing with the example in Fig.~\ref{fig:Relevance_of_KG}, we further illustrate this motivation in Fig.~\ref{fig:CLRM_example}(a). Specifically, \emph{Thunder} can be recognized as an \emph{Employer} from the semantic perspective as it is the head entity of relation \emph{employ} and is the tail entity of relation \emph{employed by}. Meanwhile, \emph{Thunder} is also a \emph{Sports team} due to the associated relation \emph{team coach}. Thus an appropriate embedding for \emph{Thunder} should be a fusion of the features representing an \emph{Employer} and a \emph{Sports team}. 
Following this intuition, an embedding method is first designed to represent entities with their relation features in original KGs. Then, the unseen entities in DEKGs can also be represented with these relation features based on this method, since the relations are shared between original KGs and DEKGs.
In this way, the seen and unseen entities can be embedded into the same feature space.
For example, if the relation-specific features for \emph{employed by}, \emph{teammate}, and \emph{coach} have been learned in the original KG, then the unseen entity \emph{Russell} in the DEKG can be directly represented with these features. Next, the relation \emph{employ} can be predicted between \emph{Thunder} and \emph{Russell} because they are recognized as an \emph{Employer} and an \emph{Employee} respectively from the semantic perspective.
Inspired by this, CLRM first extracts features for each relation and then represents entities by fusing these relation-specific features.
Formally, the set of relation-specific features $\mathcal{F}$ extracted for relations $\mathcal{R}$ is denoted as:
\begin{align}
	\mathcal{F} = \{ \boldsymbol{f}_{k} | r_k \in \mathcal{R} \},
\end{align}%
where $\boldsymbol{f}_{k} \in \mathbb{R}^{1 \times d}$ is a learned embedding we defined in our model to represent the semantics of each relation $r_k \in \mathcal{R}$, and $|\mathcal{R}|=n$.
Then the semantic information of an entity $e_i \in \mathcal{E}$ can be modeled as a relation-component table denoted as:
\begin{align}
	\mathcal{A}_i = \{ a_i^k | e_i \in \mathcal{E}, r_k \in \mathcal{R} \},
	\label{eq:table}
\end{align}%
where $a_i^k$ denotes the number of triplets with relation $r_k$ that the entity $e_i$ is associated with, and $a_i^k$ is set to $0$ if there is no triplet with relation $r_k$. 
Note that, the relation-component table of each entity is constructed using only the information of an entity's associated relations, thus Eq.~\ref{eq:table} can be generalized to both seen entities in $\mathcal{G}$ or unseen ones in $\mathcal{G}'$.
The examples of relation-component table $\mathcal{A}_i$ and $\mathcal{A}_j$ for entity $e_i$ and $e_j$ are given in Fig.~\ref{fig:Model}.

Based on the relation-component table $\mathcal{A}_i$, each entity can be represented as a fusion of corresponding relation-specific features. Formally, the semantic representation $\bold{e}_i$ of entity $e_i$ is defined as:
\begin{align}	\label{eq:fusion function}
	\bold{e}_i = \psi (\mathcal{A}_i, \mathcal{F}) = \frac{\sum_{k=0}^{n-1} a_i^k \cdot \boldsymbol{f}_k }{\sum_{k=0}^{n-1} a_i^k}, 
\end{align}%
where $\psi (\cdot)$ is the fusion function for relation-component table $\mathcal{A}_i$ and relation-specific features $\mathcal{F}$.
Notably, $\mathcal{A}_i$ is constructed from the associated triplets of entity $e_i$, and $\mathcal{F}$ is extracted for the relations shared between original KGs and DEKGs.
In this way, the representation for an entity can be calculated with only associated relations and $\mathcal{F}$, instead of initializing an embedding and fine-tuning it during training. Based on this method, CLRM can model data in an entity-independent manner and naturally generalize to unseen entities.

Finally, the score function for the semantic likelihood of a triplet $(e_i, r_k, e_j)$ is defined as: 
\begin{align}
	\label{eq:sem_score}
	\phi^{sem} (e_i, r_k, e_j) = \langle \bold{e}_i , \boldsymbol{r}_k^{sem},  \bold{e}_j \rangle, 
\end{align} where $\boldsymbol{r}_k^{sem} \in \mathbb{R}^{1 \times d}$ is a learned embedding of relation $r_k$ from the semantic perspective, $\langle , , \rangle$ denotes the element-wise product for embedding vectors inspired by DistMult \cite{DistMult}. Notably, $\boldsymbol{r}_k^{sem}$ is used as a weight matrix for the relation $r_k$ in the DistMult-based decoder to calculate the score for a triplet $(e_i, r_k, e_j)$. The reason for choosing DistMult as our decoder is that DistMult is a semantic matching model, which aligns with the intuition of CLRM that we want to extract semantic information behind KGs. What's more, although DistMult is a transductive KGE method, it is only used as a decoder thus whether it is transductive or inductive does not matter.

\subsubsection{Semantic-aware Contrastive Learning}
\label{sec:clss}
Inspired by the success of contrastive learning in graph representation learning \cite{MoCL,GCC}, a carefully designed contrastive learning based method is used to optimize the relation-specific features. The major novelty is that a semantic-aware sampling strategy is designed in the contrastive learning process.
Intuitively, the essential semantic of an entity is stable if no new relation is attached or no relation is completely removed from this entity. In other words, we assume that there is a significant change of an entity's semantic if the entity is attached with new relations or all triplets with a particular relation of the entity are deleted, which aligns with the intuition.
To model the semantic variation of entities, for each entity $e_i$, we define three different random operations, i.e., relation variation $o_1(\cdot)$,  relation addition $o_2(\cdot)$, and relation deletion $o_3(\cdot)$ for its corresponding relation-component table. 

Following the example in Fig.~\ref{fig:CLRM_example}, the social image of \emph{Russell} as a \emph{Sports player} will retain stable if he has more or fewer triplets of relation \emph{teammate} (i.e., in Fig.~\ref{fig:CLRM_example}(b), we add a triplet of \emph{teammate} with the entity $e_8$). Because there still exists triplets that can provide the semantics of \emph{Sports player} to him in the KG. But his social image will change significantly if all the triplets with relations \emph{teammate} and \emph{coach} are deleted or he is added with a new triplet of relation \emph{father of} (i.e., we delete all triplets with relations \emph{teammate} and \emph{coach} in Fig.~\ref{fig:CLRM_example}(c) and add a triplet with a new relation \emph{father of} in Fig.~\ref{fig:CLRM_example}(d)). Because the triplets that can provide him with the semantics of \emph{Sports player} are all deleted, and the added triplet with relation \emph{father of} will attach the new semantics of \emph{Father} to him.
Following this intuition, our sample strategy in the contrastive learning method generates positive examples with relation variation and negative examples with relation addition and deletion. Formally, the definitions of the three random operations are as follows.

\paragraph{Relation Variation} In operation $o_1(\cdot)$, the number of triplets with a particular relation of entity $e_i$ is randomly varied. Formally, select a number $a_i^k$ in $\{ a_i^k | a_i^k \in \mathcal{A}_i \land a_i^k \neq 0\}$, it is randomly varied to another integer in the range of $[1, m_i * \theta]$, where $\theta$ is the hyper-parameter of scaling factor, $m_i$ denotes the average number of triplets associated with each relation, denoted as:
\begin{align}
	m_i = \frac{\sum_{k=0}^{n-1} a_i^k}{| \{ a_i^k | a_i^k \in \mathcal{A}_i \land a_i^k \neq 0\} |},
\end{align}%

\paragraph{Relation Addition} In operation $o_2(\cdot)$, the triplets of a randomly selected new relation are attached to entity $e_i$.
Formally, select a number $a_i^k$ in $\{ a_i^k | a_i^k \in \mathcal{A}_i \land a_i^k = 0\}$, $a_i^k$ is randomly set to an integer in the range of $[1, m_i * \theta]$, where $m_i$ and $\theta$ are the same as $o_1(\cdot)$.

\paragraph{Relation Deletion} In operation $o_3(\cdot)$, all the triplets with a particular relation of entity $e_i$ are deleted. Formally, select a number $a_i^k$ in $\{ a_i^k | a_i^k \in \mathcal{A}_i \land a_i^k \neq 0\}$, $a_i^k$ is set to 0.

Based on the three random operations, given an entity $e_i$, a sequence of $o_1(\cdot)$ are applied to generate a positive example $e_i^{pos}$ with corresponding relation-component table $\mathcal{A}_i^{pos}$, a sequence of $o_2(\cdot)$ and $o_3(\cdot)$ are applied to generate a negative example $e_i^{neg}$ with corresponding relation-component table $\mathcal{A}_i^{neg}$. Then the representations of the positive and negative examples can be obtained by the fusion function in Eq.~(\ref{eq:fusion function}) as: 
\begin{align}
		\bold{e}_i^{pos} = \psi (\mathcal{A}_i^{pos}, \mathcal{F}), \
		\bold{e}_i^{neg} = \psi (\mathcal{A}_i^{neg}, \mathcal{F}),
\end{align}%

Next, the contrastive learning loss is calculated with a triplet loss function to maximize the similarity between positive pair $(\bold{e}_i^{pos}, \bold{e}_i)$ and minimize the similarity between negative pair $(\bold{e}_i^{neg}, \bold{e}_i)$. Formally, 
\begin{align}
\label{eq:contrastive_learning_loss}
	\mathcal{L}_c = [sim(\bold{e}_i^{pos}, \bold{e}_i) - sim(\bold{e}_i^{neg}, \bold{e}_i) + \gamma ]_+,
\end{align}%
where $[x]_+ = max\{0, x\}$ and $\gamma$ is the hyper-parameter for the margin. $sim(\cdot)$ is a function that measures the similarity between two embedding vectors by calculating the euclidean distance between them. The contrastive learning loss will be used in the final learning objective in Eq.~(\ref{eq:learning_object}) to optimize the relation-specific features. Notably, the contrastive learning is only employed during training, i.e., the above operations only consider the original KG $\mathcal{G}$.

\begin{figure}
	\centering
	\includegraphics[width=\linewidth]{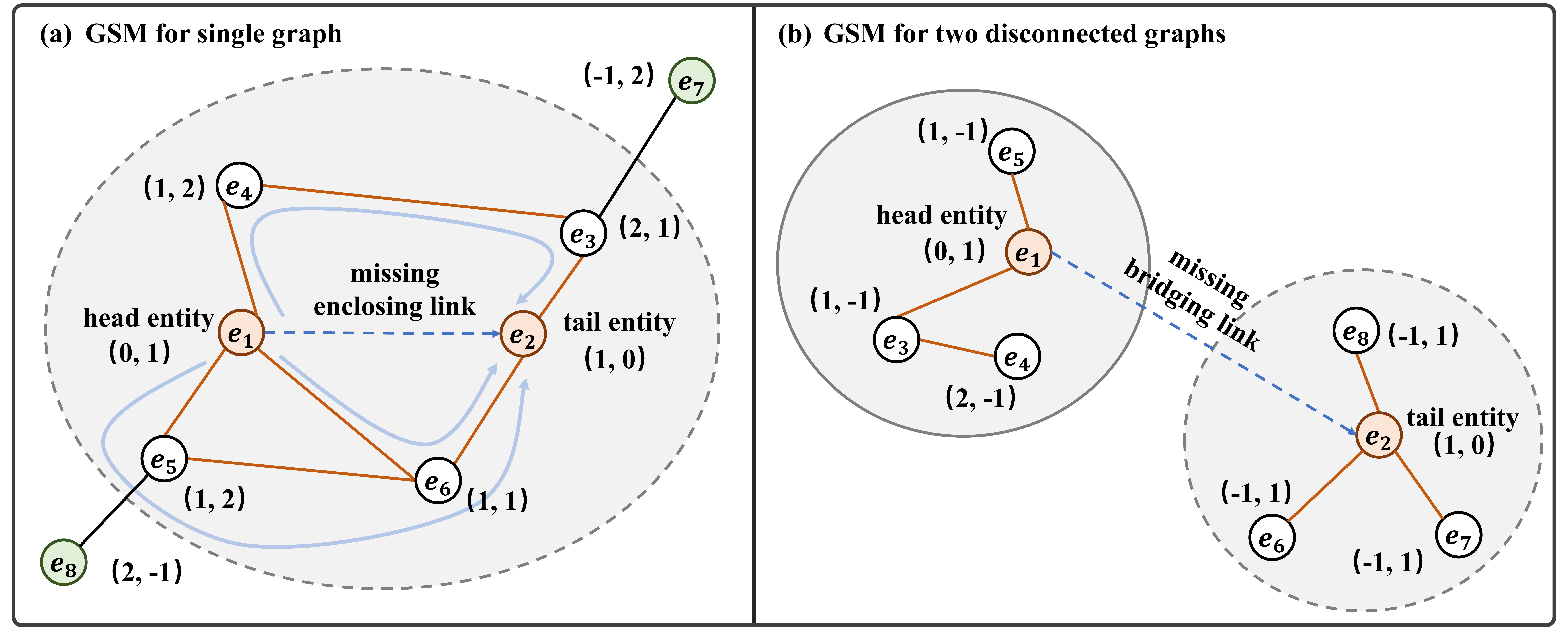}
	\vspace{-15pt}
	\caption{An example of GSM to illustrate how this module embeds a single graph for enclosing links or two disconnected graphs for bridging links.}
	\label{fig:GSM_example}
	\vspace{-15pt}
\end{figure}

\subsection{Module GSM}
To take full advantage of the topological information in the subgraph around links and relieve the \emph{topology limitation} in bridging link prediction, we extend the idea of Grail \cite{Grail} with an improved node labeling method. To be specific, the underlying idea of performing subgraph reasoning depending on the edges between entities (i.e., the solid blue arrows in Fig.~\ref{fig:GSM_example}(a)) still works in GSM when predicting \emph{enclosing links}. Then, the improved node labeling method enables GSM to extract topological information for two disconnected graphs when predicting \emph{bridging links} as shown in Fig.~\ref{fig:GSM_example}(b).

\subsubsection{Subgraph Extraction} For a triplet $(e_i, r_k, e_j)$, the subgraph $\mathcal{G}(e_i, r_k, e_j)$ is constructed based on the $t$-hop neighbors of $e_i$ and $e_j$ in $\mathcal{N}_t(e_i) \cap \mathcal{N}_t(e_j)$. The time complexity of the subgraph extraction is $\mathcal{O}(log(\mathcal{V})\mathcal{E} + \mathcal{R}dk$) \cite{Grail}, where $\mathcal{V}$, $\mathcal{R}$, and $\mathcal{E}$ denote the number of nodes, relations, and edges respectively, $k$ means extracting a $k$-hop subgraph, and $d$ denotes the dimension of embedding vectors. Note that, GSM will extract a single subgraph for an enclosing link and two disconnected subgraphs for a bridging link as shown in Fig.~\ref{fig:GSM_example}.

\subsubsection{Node Labeling} To model the relative position information in $\mathcal{G}(e_i, r_k, e_j)$, each node $e_u$ in this subgraph is labeled as $(d(i, u), d(j, u))$, where $d(i, u)$ denotes the distance of the shortest path between $e_i$ and $e_u$ without any path through $e_j$. $e_i$ and $e_j$ are uniquely labeled as $(0, 1)$ and $(1, 0)$ respectively. Then, the input embedding of node $e_u$ can be represented as $[one\verb|-|hot(d(i, u)) \oplus one \verb|-| hot(d(j, u))]$, where $\oplus$ denotes the concatenation of two embedding vectors. $one\verb|-|hot(p) \in \mathcal{R}^{1 \times d}$ represents a $d$-dimension one-hot vector, where the $p$-th entry is set to 1. 
However, the node labeling method in Grail suffers from \emph{topological limitation} and can not handle the situation in Fig.~\ref{fig:GSM_example}(b). In GSM, we improve the labeling method as follow.

Observed from Fig.~\ref{fig:GSM_example}(b), the white nodes in one subgraph is unreachable to the nodes in the other subgraph.
Grail consider only the situation in Fig.~\ref{fig:GSM_example}(a) and prunes the nodes in $\{ e_u | d(i, u) > t \lor d(j, u)>t \}$ (i.e., the green nodes) as it assumes that these nodes are redundant to form an enclosing subgraph around the target link.
However, we argue that these nodes can simulate the disconnected nodes as they are reachable to the node on one end of the target link while unreachable to the other in $t$-hop, thus still carry useful information.
Therefore, we remain the nodes in $\{ e_u | d(i, u) > t \lor d(j, u)>t \}$ and set $d(\cdot, u)$ as $-1$ if $d(\cdot, u) > t$. The one-hot vector of $one\verb|-|hot(-1)$ is set to all zero.

\subsubsection{Topological Information Modeling}
With the enclosing subgraph around link $(e_i, r_k, e_j)$ and the initial one-hot embeddings of nodes, R-GCN \cite{R-GCN} with specially designed edge attention AGGREGATE function \cite{Grail} is employed to obtain the topological information of the subgraph around the link. In particular, the architecture for the $l$-th layer of the GNN is denoted as:
\begin{align}
	& \boldsymbol{a}_i^l = {\rm{AGGREGATE}}^l ( \{ \boldsymbol{h}_s^{l-1} | s \in \mathcal{N}(i) \}, \boldsymbol{h}_i^{l-1} ), \\
	& \boldsymbol{h}_i^{l} = {\rm{COMBINE}}^l ( \boldsymbol{h}_i^{l-1}, \boldsymbol{a}_i^l), 
\end{align}%
where $\mathcal{N}(i)$ is the collection of direct neighbors of entity $e_i$, $\boldsymbol{a}_i^l$ is the aggregated message from these neighbors and $\boldsymbol{h}_i^{l}$ is the topological representation of entity $e_i$ in the $l$-th layer.
Then a $L$-layers GNN is used to obtain the representation of each entity in the subgraph $\mathcal{G}(e_i, r_k, e_j)$. The representation for the entire subgraph is obtained by applying an average-pooling on the representations of all entities as:
\begin{align}
\label{eq:GNN}
	& \boldsymbol{h}_{\mathcal{G}(e_i, r_k, e_j)}^L = \frac{1}{\mathcal{V}} \sum_{v \in \mathcal{V}} \boldsymbol{h}_v^L,
\end{align}%
where $\mathcal{V}$ denotes the set of nodes in subgraph $\mathcal{G}(e_i, r_k, e_j)$.
Finally, the score for the topological likelihood of the link $(e_i, r_k, e_j)$ is given by:
\begin{align}
	\phi^{tpo} (e_i, r_k, e_j) \!= \! [\boldsymbol{h}_{\mathcal{G}(e_i, r_k, e_j)}^L \oplus \boldsymbol{h}_i^{L} \oplus \boldsymbol{h}_j^{L} \oplus \boldsymbol{r}_k^{tpo}] \boldsymbol{W},
	\label{eq:tpo_score}
\end{align}%
where $\boldsymbol{r}_k^{tpo} \in \mathbb{R}^{1 \times d}$ is a learned embedding of relation $r_k$ from the topological perspective and $\boldsymbol{W}$ is a linear weight matrix.

\subsection{Training Objective}
To train the entire model, all triplets in the original KG are naturally served as positive triplets denoted as $\mathcal{T}_{pos}$. Then the negative sampling is performed on each $(e_i, r_k, e_j) \in \mathcal{T}_{pos}$ by randomly corrupting the head or tail entity with another entity in $\mathcal{E}$ to construct negative triplet set $\mathcal{T}_{neg}$. Formally, the set of corrupted negative triplets is denoted as:
\begin{align}
	\mathcal{T}_{neg} = \{ (e_i', r_k, e_j) | e_i' \in \mathcal{E} \} \cup \{ (e_i, r_k, e_j') | e_j' \in \mathcal{E} \}.
\end{align}%

To encourage the decoder to consider both the global semantic information and local topological information, the score of each link is calculated as the sum of scores in Eq.~(\ref{eq:sem_score}) and Eq.~(\ref{eq:tpo_score}). Formally, the score is defined as:
\begin{align}
	\label{eq:final_score}
	\phi (e_i, r_k, e_j) = \phi^{sem} (e_i, r_k, e_j) + \phi^{tpo} (e_i, r_k, e_j).
\end{align}%
Then we employ a margin-based ranking loss $\mathcal{L}_s (e_i, r_k, e_j)$ to assign high scores for positive triplets and low scores for negative ones,
\begin{align}
\label{eq:triplet_score_loss}
	\mathcal{L}_s (e_i, r_k, e_j) = [\gamma - \phi(e_i, r_k, e_j) + \phi(e_i', r_k, e_j')]_+.
\end{align}%
Finally, the overall training objective of the proposed model DEKG-ILP is to minimize the final loss $\mathcal{L}_f$ that is defined as:
\begin{align}
	\label{eq:learning_object}
	\mathcal{L}_f = \sum_{\mathcal{T}_{pos}} \sum_{\mathcal{T}_{neg}} \mathcal{L}_s + \sigma \sum_{\mathcal{T}_{pos}} \mathcal{L}_c,
\end{align}%
where $\sigma$ is a hyper-parameter. 
Notably, the contrastive learning loss $\mathcal{L}_c$ is only used during training to optimize the relation-specific features. The score for each link during testing is directly calculated by the score function in Eq.~(\ref{eq:final_score}). The training process of DEKG-ILP is summarized in Algorithm~\ref{alg:training}.

\begin{algorithm}[t]
	\caption{Training process of DEKG-ILP}
	\label{alg:training}
	\begin{algorithmic}[1]
		\REQUIRE ~~\\
		The positive triplets $\mathcal{T}_{pos}$ and negative triplets
		$\mathcal{T}_{neg}$; \\
	    Subgraph $\mathcal{G}_{(e_i, r_k, e_j)}$ for each link; \\
		Relation-component table $\mathcal{A}_i$ for each entity $e_i$. \\
		Positive and negative examples $\mathcal{A}_i^{pos}$ and $\mathcal{A}_i^{neg}$ for each entity $e_i$.;
		\STATE \textbf{Initialize:} \\
		\STATE The relation-specific features $\mathcal{F}$; \\
		\STATE The learned parameters $\Theta$ in GNNs model; \\
		\STATE The learned embeddings $\boldsymbol{r}_k^{sem}$ and $\boldsymbol{r}_k^{tpo}$; \\
		
		\REPEAT
		    \FOR{each triplet ($e_i, r_k, e_j)$ in the batch}
		    \STATE Get the embedding for entity $e_i$ and $e_j$ with $\mathcal{A}_i$, $\mathcal{A}_j$, $\mathcal{F}$ in Eq.~(\ref{eq:fusion function}).
		    \STATE Calculate the score $\phi^{sem} (e_i, r_k, e_j)$ in Eq.~(\ref{eq:sem_score});
		    \STATE Input the subgraph $\mathcal{G}_{(e_i, r_k, e_j)}$ into the GNN model and get $\boldsymbol{h}_{\mathcal{G}(e_i, r_k, e_j)}^L, \boldsymbol{h}_i^{L}, \boldsymbol{h}_j^{L}$ at the $L$ layer;
		    \STATE Get topological score $\phi^{tpo} (e_i, r_k, e_j)$ in Eq.~(\ref{eq:tpo_score});
		    \STATE Combine $\phi^{sem}$ and $\phi^{tpo}$ to obtain $\phi$ in Eq.~(\ref{eq:final_score});
		    \STATE Calculate the socres for the corresponding negative triplets in the same way.
		    \STATE Calculate the triplet score loss $\mathcal{L}_s$ in Eq.~(\ref{eq:triplet_score_loss});
		    \STATE Get the embedding $\bold{e}_i^{pos}$ and $\bold{e}_i^{neg}$ for the positive and negative examples with $\mathcal{A}_i^{pos}$, $\mathcal{A}_i^{neg}$, and $\mathcal{F}$;
		    \STATE Compute the contrastive loss $\mathcal{L}_c$ in Eq.~(\ref{eq:contrastive_learning_loss});
		    \STATE Minimize the final loss $\mathcal{L}_f$ in Eq.~(\ref{eq:learning_object}) and update $\mathcal{F}, \Theta, \boldsymbol{r}_k^{sem}, \boldsymbol{r}_k^{tpo}$ in the back-propagation process;
		    \ENDFOR
		\UNTIL {the final epoch}
	\ENSURE $\mathcal{F}, \Theta, \boldsymbol{r}_k^{sem}, \boldsymbol{r}_k^{tpo}.$
	\end{algorithmic}
\end{algorithm}

\section{Experiments}
\label{sec:experiments}

In this section, we first introduce the experimental configuration, including datasets, baselines, evaluation metrics, and parameter setup. Then, we present the main results of our proposed model and all compared baselines on several benchmark datasets. We further provide the performance of DEKG-ILP on predicting \emph{enclosing links} only and \emph{bridging links} only respectively. Moreover, ablation study, complexity analysis and case study are also presented. The source code is available at \url{https://github.com/Ninecl/DEKG-ILP}.

\subsection{Dataset}
To evaluate the performance of all compared methods on inductive link prediction with both \emph{enclosing links} and \emph{bridging links} in DEKGs, additional links are extracted from corresponding real-world raw KGs for testing based on the benchmark datasets provided by Grail \cite{Grail}. The final evaluation datasets are constructed by mixing up \emph{enclosing links} and \emph{bridging links} in different ratios to consider the impact of different data compositions.

\begin{table}[]
	\centering
	\caption{Statistics of datasets, $\mathcal{G}$ and $|\mathcal{R}|$, $|\mathcal{E}|$, and $|\mathcal{T}|$ denote the numbers of relations, entities, and triplets in $\mathcal{G}$ and $\mathcal{G}'$}
	\label{tab:datasets}
    \resizebox{\linewidth}{!}{
		\setlength{\tabcolsep}{1mm}{
		    \begin{tabular}{ccccccccccc}
                \toprule
                \multicolumn{2}{c}{\multirow{2}{*}{}} & \multicolumn{3}{c}{FB15k-237} & \multicolumn{3}{c}{NELL-995} & \multicolumn{3}{c}{WN18RR} \\
                \multicolumn{2}{c}{} & \multicolumn{1}{c}{$|\mathcal{R}|$} & \multicolumn{1}{c}{$|\mathcal{E}|$} & $|\mathcal{T}|$ & \multicolumn{1}{c}{$|\mathcal{R}|$} & \multicolumn{1}{c}{$|\mathcal{E}|$} & $|\mathcal{T}|$ & \multicolumn{1}{c}{$|\mathcal{R}|$}  & \multicolumn{1}{c}{$|\mathcal{E}|$} & $|\mathcal{T}|$ \\ 
                \midrule
                
                \multicolumn{1}{c}{\multirow{2}{*}{\emph{EQ}}} & $\mathcal{G}$ & \multicolumn{1}{c}{180} & \multicolumn{1}{c}{1594} & 5226 & \multicolumn{1}{c}{14} & \multicolumn{1}{c}{3103} & 5540 & \multicolumn{1}{c}{9} & \multicolumn{1}{c}{2746} & 6678 \\ 
                \multicolumn{1}{c}{} & $\mathcal{G}'$ & \multicolumn{1}{c}{142} & \multicolumn{1}{c}{1093} & 2404 & \multicolumn{1}{c}{14} & \multicolumn{1}{c}{225} & 1034 & \multicolumn{1}{c}{8} & \multicolumn{1}{c}{922} & 1991 \\ 
                \midrule
                
                \multicolumn{1}{c}{\multirow{2}{*}{\emph{MB}}} & $\mathcal{G}$  & \multicolumn{1}{c}{200} & \multicolumn{1}{c}{2608} & 12085 & \multicolumn{1}{c}{88}  & \multicolumn{1}{c}{2564} & 10109 & \multicolumn{1}{c}{10} & \multicolumn{1}{c}{6954} & 18968 \\ 
                \multicolumn{1}{c}{} & $\mathcal{G}'$ & \multicolumn{1}{c}{172} & \multicolumn{1}{c}{1660} & 5570 & \multicolumn{1}{c}{79} & \multicolumn{1}{c}{2086} & 5997 & \multicolumn{1}{c}{10} & \multicolumn{1}{c}{2757} & 5304 \\ 
                \midrule
                
                \multicolumn{1}{c}{\multirow{2}{*}{\emph{ME}}} & $\mathcal{G}$  & \multicolumn{1}{c}{215} & \multicolumn{1}{c}{3668} & 22394 & \multicolumn{1}{c}{142} & \multicolumn{1}{c}{4647} & 20117 & \multicolumn{1}{c}{11} & \multicolumn{1}{c}{12078} & 32150 \\
                \multicolumn{1}{c}{} & $\mathcal{G}'$ & \multicolumn{1}{c}{183} & \multicolumn{1}{c}{2501} & 9569 & \multicolumn{1}{c}{122} & \multicolumn{1}{c}{3566} & 10072 & \multicolumn{1}{c}{11} & \multicolumn{1}{c}{5084} & 7772 \\ 
                
                \bottomrule
            \end{tabular}
        }
    }
    \vspace{-10pt}
\end{table}

Specifically, Grail has extracted four different datasets v1, v2, v3, and v4 from three raw real-world KGs (i.e., FB15k-237 \cite{ConvE}, NELL-995 \cite{NELL-995}, and WN18RR \cite{WN18RR}) respectively with different scales, and these datasets have been split into the original KG $\mathcal{G}$ for training and the DEKG $\mathcal{G}'$ for testing. In our experiments, we further construct three datasets,  \emph{EQ} (equal links), \emph{MB} (more \emph{bridging links}), \emph{ME} (more \emph{enclosing links}) for FB15k-237, NELL-995, and WN18RR respectively based on datasets v1, v2, and v3 released in Grail.
During our training stage, $\mathcal{G}$ is used as training set same as Grail. During the testing stage, apart from triplets in $\mathcal{G}'$ which serve as \emph{enclosing links} for evaluation, we extract certain number of triplets that bridge $\mathcal{G}$ and $\mathcal{G}'$ from the corresponding raw KGs as \emph{bridging links} for evaluation as well. \textbf{ Note that, these bridging links are real links extracted from the raw KGs}.
Finally, the evaluation datasets are constructed by mixing up these \emph{enclosing links} and \emph{bridging links} in the ratios of 1:1, 1:2, 2:1 for \emph{EQ}, \emph{MB}, \emph{ME} respectively. TABLE~\ref{tab:datasets} presents the statistics of these datasets.

\subsection{Baseline}
The models Grail \cite{Grail} and TACT \cite{TACT} are used as main baselines since they are both proposed for DEKGs. RuleN \cite{RuleN}, GEN \cite{GEN}, TransE \cite{TransE}, RotatE \cite{RotatE}, and ConvE \cite{ConvE} are compared as the representation of rule-mining based, GNNs based, distance based, rotation based, and neural network based methods to explore how these methods perform for DEKGs. Note that, Grail, TACT, RuleN, and GEN can be directly applied to the above constructed DEKG datasets as they are inductive inherently. We implement these four baselines according to the source codes they released online. To implement the rest transductive methods TransE, RotatE, ConvE in an inductive scenario, OpenKE \cite{openke} is extended as follow: we first train these methods on the original KG $\mathcal{G}$ to get the embeddings of seen entities and relations. Then, the embeddings of unseen entities in the emerging KG $\mathcal{G}'$ are randomly initialized because they cannot be obtained during training. Finally, we calculate the scores for inductive links with these embeddings. Notably, all baselines are implemented with the optimal parameter settings reported in their papers.

\subsection{Evaluation Metric}
Like most related studies, Mean Reciprocal Rank (MRR) and Hits at N (Hits@N) are used as the evaluation metrics in our experiments. As Grail is evaluated on head/tail prediction while TACT only considers relation prediction, for a fair comparison, we extend these baselines to all the forms of prediction tasks including $(h, r, ?)$, $(h, ?, t)$, and $(?, r, t)$. All the negative triplets for testing are constructed by replacing elements in triplets with the candidate entity and relation set containing all entities and relations in $\mathcal{G}$ and $\mathcal{G}'$. The ranks are measured in a filtered setting where all the triplets appeared in training, valid, and test set are removed.
What's more, all the models are run five times on each dataset with different random seeds and the average results are reported. 

\subsection{Parameter Setup}
The hyper-parameters of our model DEKG-ILP include learning rate $lr$ during training time, embedding dimension $d$ of relation-specific features, edge dropout rate $\beta$ that denotes the dropout percent of the edges in the GNN model, and the loss coefficient $\sigma$ used to adjust the weight of contrastive learning loss in Eq. (\ref{eq:learning_object}). A grid search is conducted on the validation sets to find the hyper-parameters with optimal performance. We sample 1 negative triplet for each positive triplet in training set to calculate the triplet score loss in Eq. (\ref{eq:triplet_score_loss}), sample 10 positive and negative examples for each entity respectively to calculate the contrastive learning loss in Eq. (\ref{eq:contrastive_learning_loss}).
We fine-tune the learning rate $lr$ in \{0.1, 0.01, 0.001, 0.0005\}, relation-specific feature dimension $d$ in \{16, 32, 64, 128\}, edge dropout $\beta$ in \{0.1, 0.3, 0.5, 0.8\}, and coefficient $\sigma$ in \{0.01, 0.1, 0.5, 1\}. The optimal configuration in the inductive link prediction task is $lr=0.01$, $d=32$, $\beta=0.5$, $\sigma=0.1$

\begin{table*}[t]
\caption{Main result on \emph{EQ}, \emph{MB}, \emph{ME} of FB15k-237, NELL-995, and WN18RR} 
\centering
\resizebox{0.98\textwidth}{!}
{
    \begin{tabular}{cccccccccccccc}
        \toprule
        \multirow{2}{*}{Datasets} & \multirow{2}{*}{Models} & \multicolumn{4}{c}{\emph{EQ}} & \multicolumn{4}{c}{\emph{MB}} & \multicolumn{3}{c}{\emph{ME}} \\ 
        \cmidrule(r){3-6} \cmidrule(r){7-10} \cmidrule(r){11-14}
        & & MRR & Hits@1 & Hits@5 & Hits@10 & MRR & Hits@1 & Hits@5 & Hits@10 & MRR & Hits@1 & Hits@5 & Hits@10 \\
        \midrule
        \multirow{8}{*}{\rotatebox{90}{FB15k-237}} 
            & TransE    & 0.241 & 0.169 & 0.264 & 0.337 & 0.210 & 0.143 & 0.249 & 0.310 & 0.197 & 0.123 & 0.218 & 0.317 \\
            & RotatE    & 0.089 & 0.021 & 0.095 & 0.192 & 0.101 & 0.025 & 0.125 & 0.233 & 0.094 & 0.022 & 0.105 & 0.209 \\
            & ConvE     & 0.102 & 0.033 & 0.105 & 0.227 & 0.097 & 0.031 & 0.105 & 0.200 & 0.119 & 0.030 & 0.114 & 0.228 \\
            & GEN       & 0.093 & 0.032 & 0.089 & 0.196 & 0.109 & 0.030 & 0.130 & 0.241 & 0.101 & 0.027 & 0.110 & 0.207 \\
            & RuleN     & 0.265 & \underline{0.237} & 0.267 & 0.268 & 0.212 & \underline{0.186} & 0.239 & 0.240 & 0.402 & 0.360 & 0.443 & 0.447 \\
            & Grail 	& \underline{0.279} & 0.216 & \underline{0.323} & 0.342 & \underline{0.226} & 0.164 & \underline{0.259} & 0.281 & \underline{0.456} & \underline{0.378} & \underline{0.523} & \underline{0.569} \\
            & TACT 		& 0.227 & 0.130 & 0.316 & \underline{0.401} & 0.186 & 0.101 & 0.249 & \underline{0.339} & 0.311 & 0.222 & 0.382 & 0.463 \\
    		\cmidrule(r){2-14}
            & DEKG-ILP	& \textbf{0.508} & \textbf{0.351} & \textbf{0.693}& \textbf{0.841} & \textbf{0.535} & \textbf{0.396} & \textbf{0.693} & \textbf{0.832} & \textbf{0.634} & \textbf{0.512} & \textbf{0.785} & \textbf{0.891} \\
        \midrule
        
        \multirow{8}{*}{\rotatebox{90}{NELL-995}} 
            & TransE 	& 0.083 & 0.023 & 0.079 & 0.156 & 0.234 & 0.161 & 0.263 & 0.345 & 0.158 & 0.088 & 0.177 & 0.269 \\
            & RotatE	& 0.118 & 0.021 & 0.170 & 0.331 & 0.090 & 0.027 & 0.089 & 0.179 & 0.091 & 0.014 & 0.094 & 0.187 \\
            & ConvE     & 0.098 & 0.035 & 0.108 & 0.172 & 0.102 & 0.033 & 0.088 & 0.194 & 0.100 & 0.026 & 0.102 & 0.179 \\
            & GEN       & 0.091 & 0.029 & 0.084 & 0.181 & 0.132 & 0.105 & 0.156 & 0.239 & 0.127 & 0.058 & 0.142 & 0.223\\
            & RuleN     & \underline{0.234} & \underline{0.197} & 0.230 & 0.258 & 0.300 & \underline{0.249} & 0.336 & 0.340 & \underline{0.452} & \underline{0.371} & 0.508 & 0.510 \\
            & Grail 	& 0.193 & 0.109 & \underline{0.231} & \underline{0.393} & \underline{0.307} & 0.233 & \underline{0.352} & \underline{0.423} & 0.411 & 0.305 & \underline{0.518} & \underline{0.614} \\
            & TACT 		& 0.156 & 0.071 & 0.221 & 0.328 & 0.223 & 0.125 & 0.305 & 0.420 & 0.292 & 0.165 & 0.428 & 0.557 \\
    		\cmidrule(r){2-14}
            & DEKG-ILP	& \textbf{0.353} & \textbf{0.218} & \textbf{0.489}& \textbf{0.631} & \textbf{0.468} & \textbf{0.301} & \textbf{0.694} & \textbf{0.830} & \textbf{0.532} & \textbf{0.380} & \textbf{0.727} & \textbf{0.842} \\
        \midrule
        
        \multirow{8}{*}{\rotatebox{90}{WN18RR}} 
            & TransE    & 0.133 & 0.073 & 0.135 & 0.353 & 0.164 & 0.105 & 0.167 & 0.369 & 0.160 & 0.065 & 0.126 & 0.389 \\
            & RotatE    & 0.161 & 0.038 & 0.271 & 0.476 & 0.142 & 0.030 & 0.215 & 0.458 & 0.135 & 0.028 & 0.197 & 0.407 \\
            & ConvE     & 0.111 & 0.051 & 0.124 & 0.308 & 0.122 & 0.049 & 0.165 & 0.333 & 0.129 & 0.049 & 0.115 & 0.327\\
            & GEN       & 0.158 & 0.101 & 0.147 & 0.372 & 0.153 & 0.098 & 0.170 & 0.365 & 0.148 & 0.051 & 0.119 & 0.352\\
            & RuleN     & 0.382 & \underline{0.342} & 0.402 & 0.410 & 0.252 & \underline{0.232} & 0.241 & 0.466 & 0.335 & \textbf{0.312} & 0.334 & 0.434 \\
            & Grail 	& 0.401 & 0.320 & 0.473 & \underline{0.613} & 0.261 & 0.179 & 0.344 & \underline{0.513} & \underline{0.341} & 0.243 & 0.424 & \underline{0.611} \\
            & TACT 		& \underline{0.442} & 0.328 & \underline{0.578} & 0.593 & \underline{0.335} & 0.231 & \underline{0.455} & 0.472 & 0.321 & 0.224 & \underline{0.431} & 0.475 \\
    		\cmidrule(r){2-14}
            & DEKG-ILP		& \textbf{0.471} & \textbf{0.350} & \textbf{0.607}& \textbf{0.701} & \textbf{0.359} & \textbf{0.240} & \textbf{0.480} & \textbf{0.625} & \textbf{0.378} & \underline{0.245} & \textbf{0.534} & \textbf{0.685} \\
        \bottomrule
    \end{tabular}
}
\vspace{-6pt}
\label{tab:main_result}
\end{table*}

\subsection{Main Results}
The overall results, where both \emph{enclosing links} and \emph{bridging links} are contained in the test set, of all methods on \emph{EQ}, \emph{MB}, \emph{ME} of FB15k-237, NELL-995, and WN18RR are presented in TABLE~\ref{tab:main_result}. MRR, Hits@1, Hits@5, and Hits@10 are reported here.
Several observations can be obtained from these tables: 
1) DEKG-ILP outperforms all baselines consistently across all datasets which benefits from a careful design that enables both enclosing and bridging link prediction. More details of how DEKG-ILP performs for \emph{enclosing links} only and \emph{bridging links} only are discussed in Section~\ref{sec:respective_study}.
2) The improvements of DEKG-ILP on FB15k-237 and NELL-995 are more obvious than those on WN18RR.
This phenomenon may be caused by the different number of relations in the datasets as presented in TABLE~\ref{tab:datasets}, where the number $|\mathcal{R}|$ in FB15k-237 and NELL-995 is much larger than that in WN18RR, demonstrating that the proposed model can extract richer semantic information if the given KG contains more relations. 
3) The improvement of DEKG-ILP on \emph{MB} is more obvious than that on \emph{EQ} and \emph{ME}, as more \emph{bridging links} are contained by \emph{MB}. This observation combined with the first observation shows that while our model achieves performance improvements compared with existing works in predicting both \emph{enclosing links} and \emph{bridging links}, more improvements come from the prediction of \emph{bridging links} compared with \emph{enclosing links}. The underlying reason is that our model is able to do both enclosing link and bridging link prediction while the existing works can only predict \emph{enclosing links}.
4) Grail performs better than other baselines on most datasets as it was specially designed for DEKGs. The subgraph reasoning method enable Grail to have a good performance on predicting \emph{enclosing links} in DEKGs. However, Grail suffers from the \emph{topological limitation} and cannot handle bridging link prediction task, thus it still performs worse than DEKG-ILP.
5) Although built from Grail, TACT does not perform as well as Grail, since TACT specially considers six different topological interaction of relations and achieves a good performance on relation prediction but performs poorly on head and tail prediction, thus underperforms Grail and DEKG-ILP in general.
6) RuleN achieves good results at Hits@1 on most datasets especially on \emph{ME} of WN18RR, but cannot maintain the same performance on Hits@5 and Hits@10. This is because it only focuses on whether the rule paths exist or not (i.e., 1 or 0), instead of calculating the probability of a rule path. 
7) Although GEN is also an inductive method, it does not achieve good performance because GEN tries to embed unseen entities by transforming information from seen entities to unseen ones through the edges between them, which do not exist between the original KG and DEKG. Thus GEN cannot embed and predict links for the unseen entities in DEKGs effectively.
8) The rest three transductive methods TransE, RoateE, and ConvE all have a poor performance on these datasets, demonstrating that the transductive methods are not suitable for the inductive scenario, even though a more complex model is used as the encoder in these methods.

\subsection{Respective Study}
\label{sec:respective_study}
\begin{figure}
	\centering
	\subfigbottomskip=-2pt
	\subfigure[FB15k-237 enclosing]
	{
		\centering
		\includegraphics[width=0.45\linewidth]{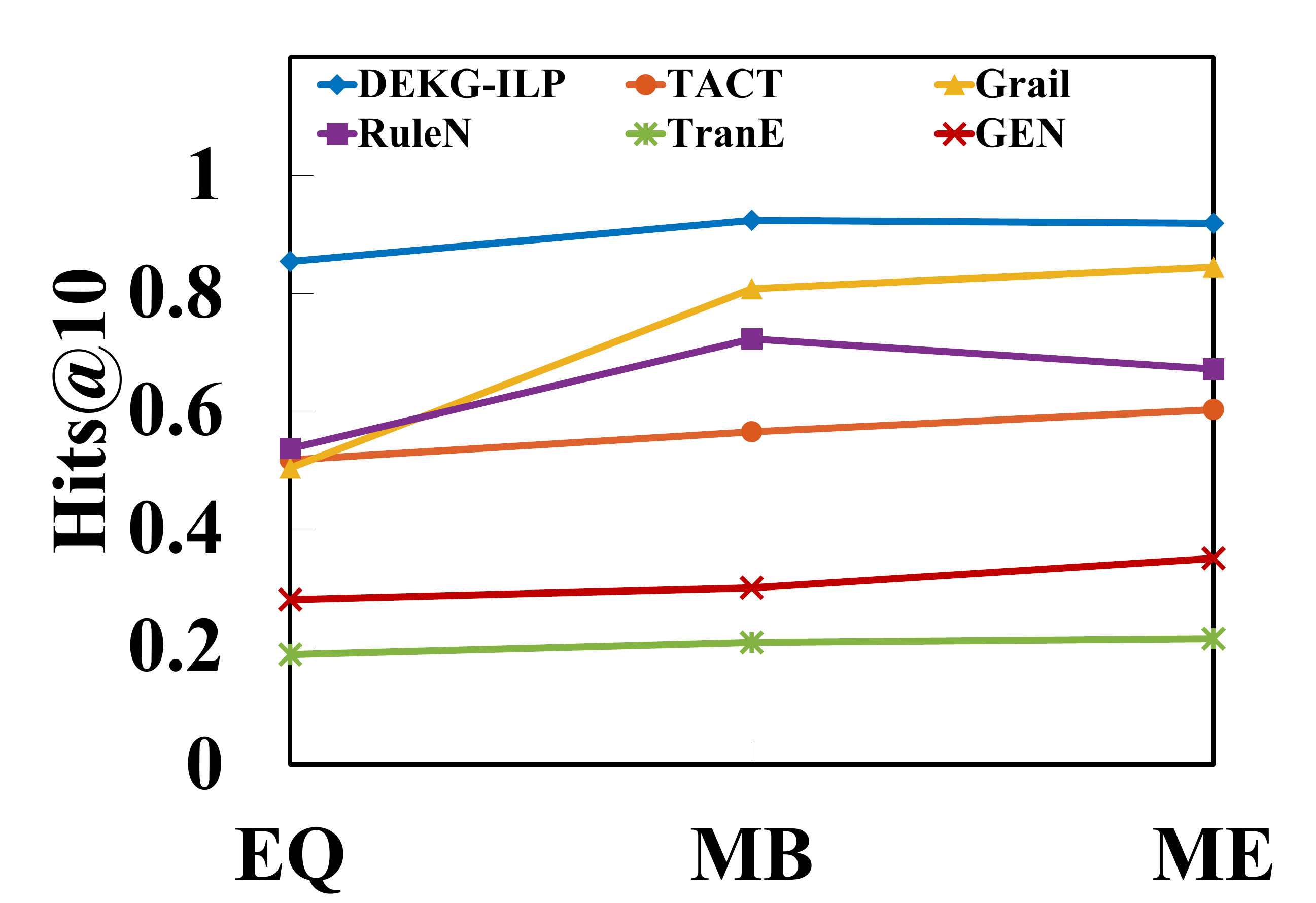}
	}
	\subfigure[FB15k-237 bridging]
	{
		\centering
		\includegraphics[width=0.45\linewidth]{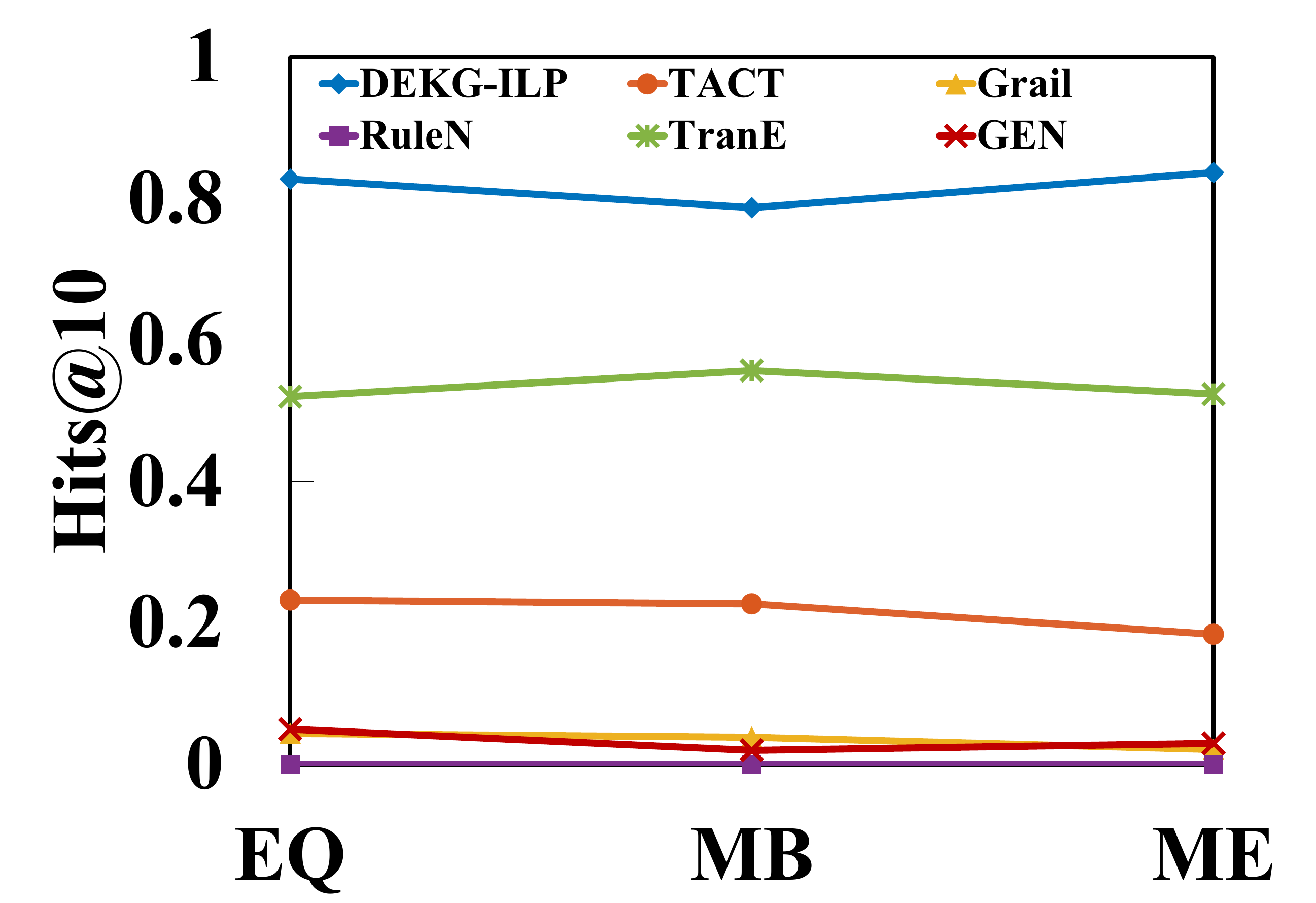}
	}
	\subfigure[NELL-995 enclosing]
	{
		\centering
		\includegraphics[width=0.45\linewidth]{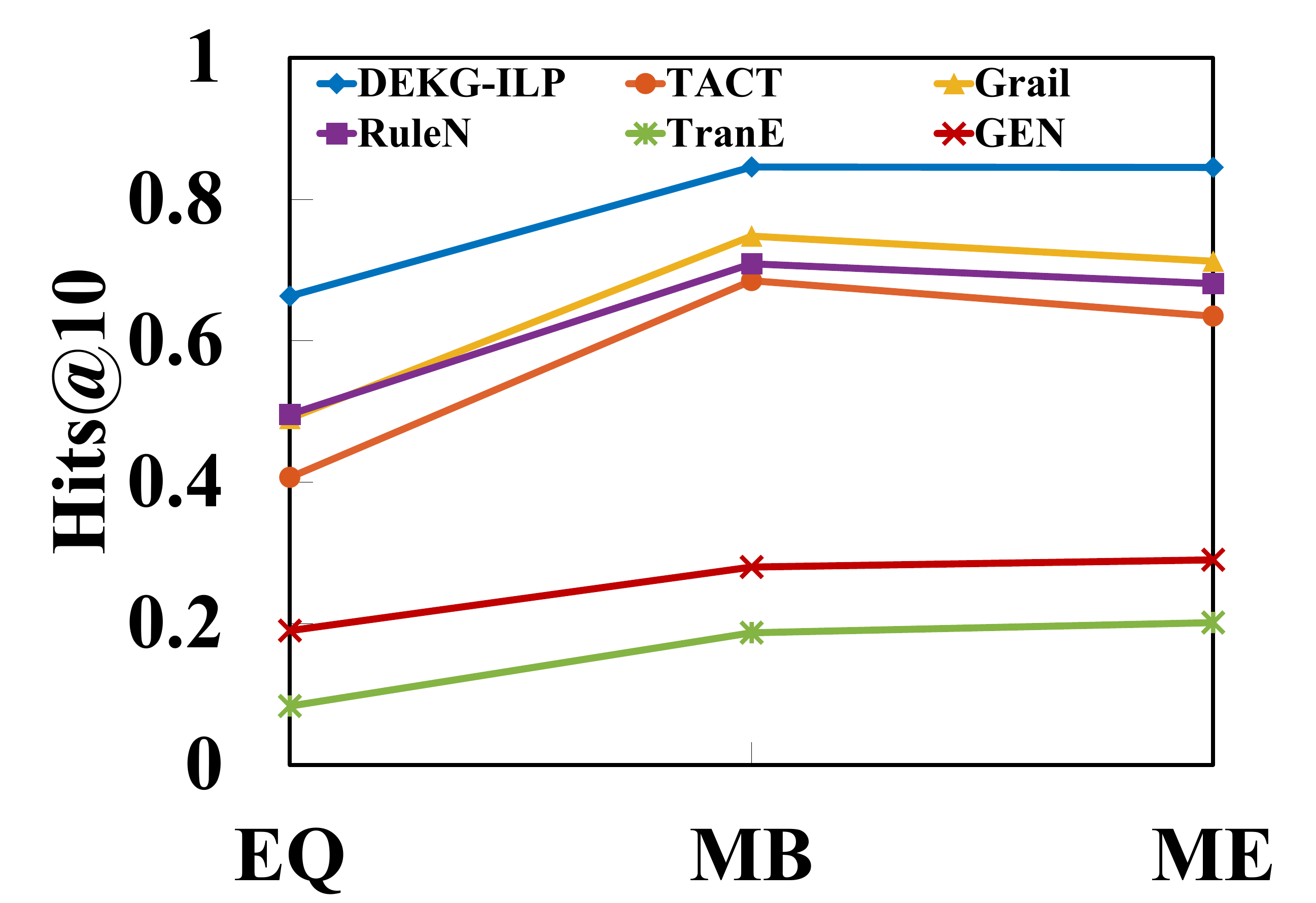}
	} 
	\subfigure[NELL-995 bridging]
	{
		\centering
		\includegraphics[width=0.45\linewidth]{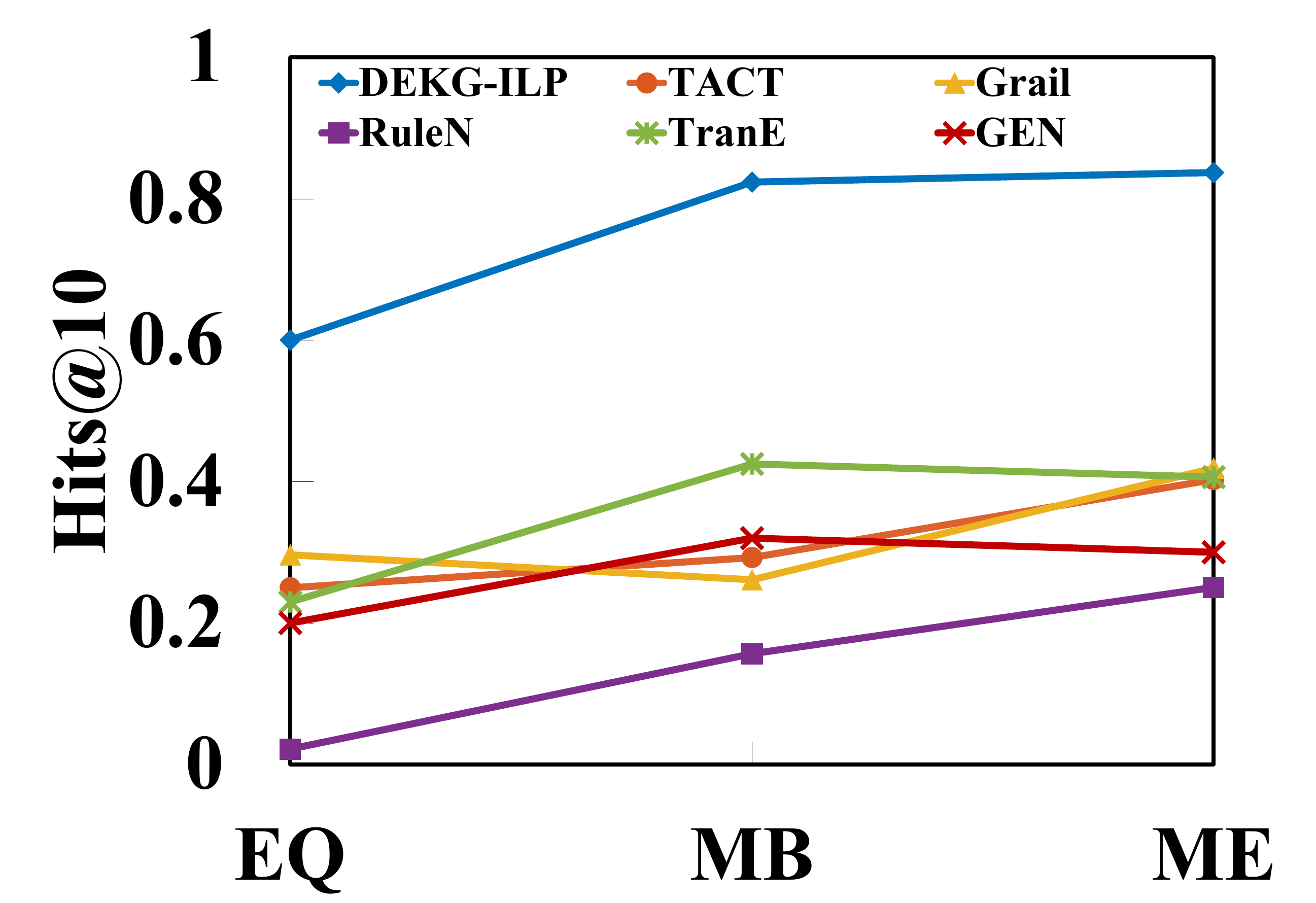}
	}
	\subfigure[WN18RR enclosing]
	{
		\centering
		\includegraphics[width=0.45\linewidth]{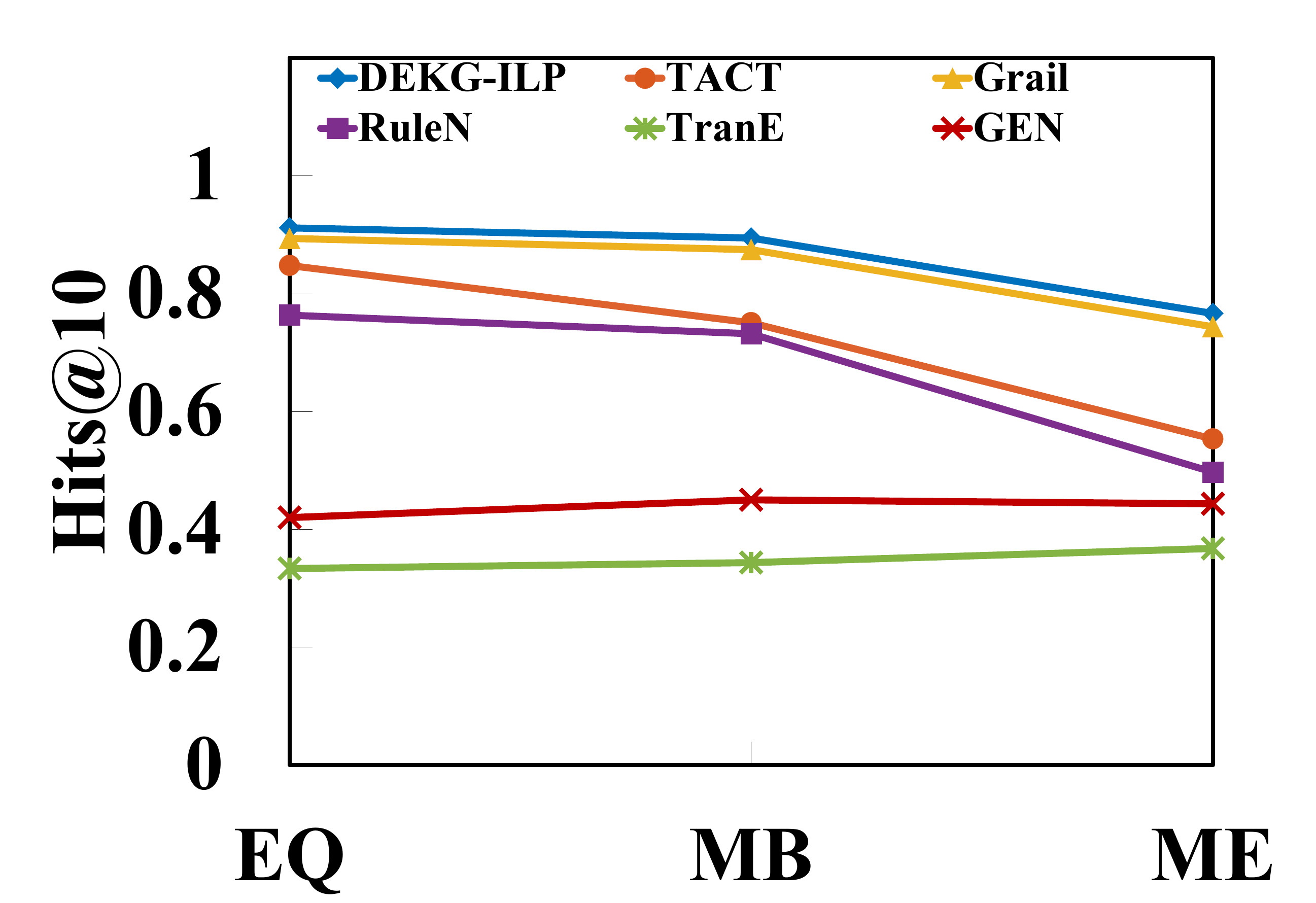}
	} 
	\subfigure[WN18RR bridging]
	{
		\centering
		\includegraphics[width=0.45\linewidth]{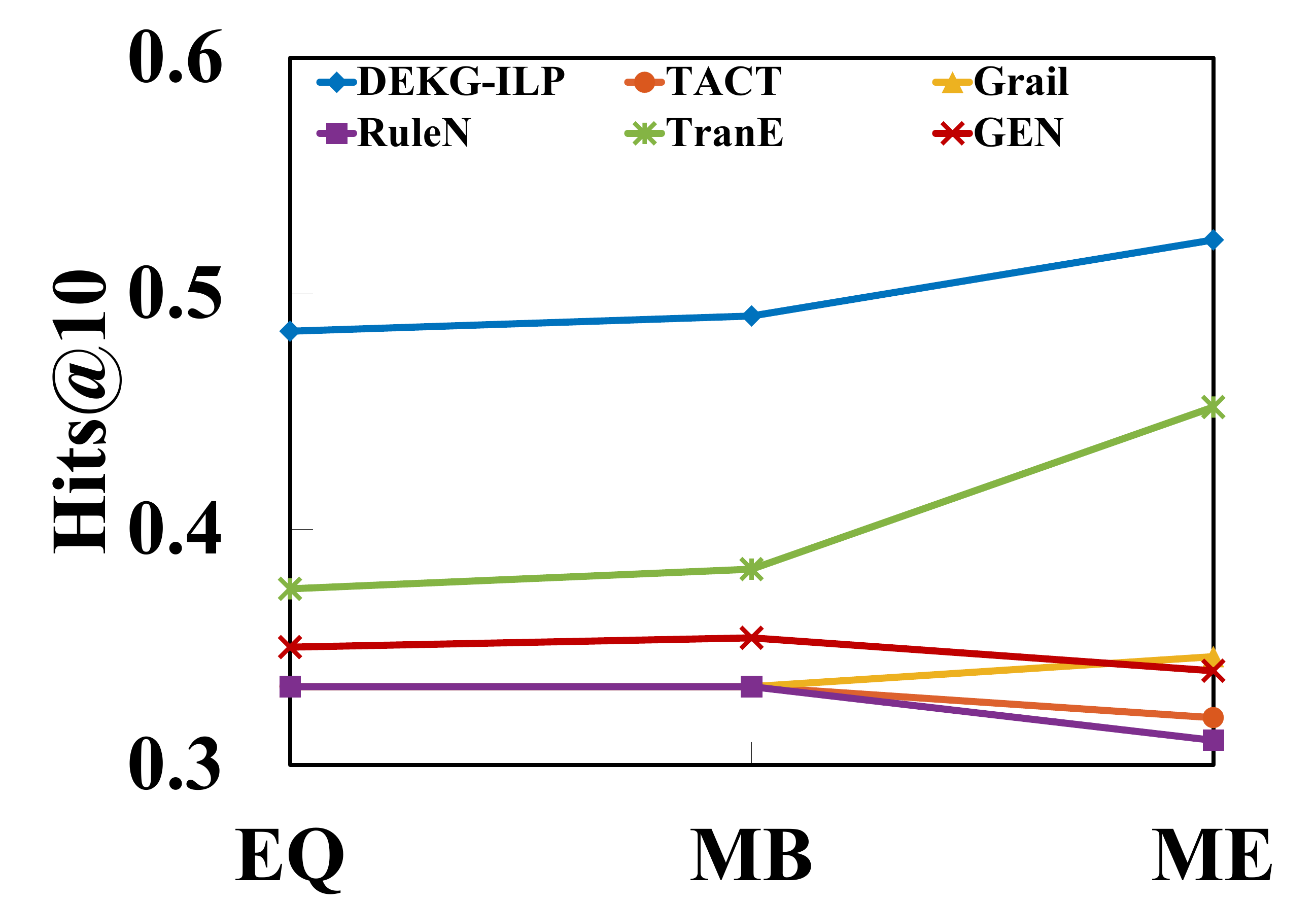}
	}
	\caption{Results of enclosing and bridging link prediction task respectively on Hits@10}
	\label{fig:respective}
	\vspace{-10pt}
\end{figure}

In this section, we evaluate models for \emph{enclosing links} only and \emph{bridging links} only respectively. Fig.~\ref{fig:respective} presents the Hits@10 of methods on datasets \emph{EQ}, \emph{MB}, and \emph{ME} with either \emph{enclosing links} only or \emph{bridging links}, where MRR, Hits@1, and Hits@5 are omitted due to the space limitation. 
It can be observed that: 
1) DEKG-ILP consistently outperforms all baselines on both enclosing and bridging link prediction tasks across all datasets. The impressive gap between DEKG-ILP and other baselines when predicting \emph{bridging links} demonstrates that our proposed model can handle the bridging link prediction task ignored by previous work, benefiting from the global semantic information extracted in CLRM. Additionally, the improvement of DEKG-ILP on \emph{enclosing links} demonstrate that our model also have a better performance on predicting \emph{enclosing links} compared with the baseline methods.
2) Although both TACT and Grail perform well on \emph{enclosing links}, our model DEKG-ILP still performs better than them. The possible reason is that they only exploit the local topological information in KGs while DEKG-ILP also exploits global relation-based semantic information. Furthermore, the reason behind their poor performance on \emph{bridging links} is that the subgraph reasoning in Grail and TACT is seriously dependent on the enclosing subgraph structure, which is missing between two disconnected KGs.
3) TransE gives very limited performance in predicting \emph{enclosing links} because all entities are unseen during testing. However, it is able to predict \emph{bridging links} to some extent. The possible reason is that the distance translation in embedding space proposed in TransE can capture the relevance between original KGs and DEKGs. Note that, we choose TransE as the representation of the three transductive methods and do not report the results of ConvE and RotatE since TransE achieve the best performance among the three transductive methods.
4) RuleN is able to predict \emph{enclosing links} to some extent by mining logical-rules. However, it shows very limited performance in predicting \emph{bridging links} because rule-mining methods seriously depend on the observed edges between entities, which are missing between original KGs and DEKGs.
5) GEN has a poor performance on both \emph{enclosing links} and \emph{bridging links} because it embeds unseen entities depending on the edges between seen entities and unseen ones, which do not exist in our scenario, and the final embeddings of unseen entities in GEN are close to random initialized vectors, similar to that in TransE.

\subsection{Ablation Study}
\label{sec:ablation}
\begin{figure*}
	\centering
	\subfigbottomskip=-1pt
	\subfigure[FB15k-237 EQ]
	{
		\centering
		\includegraphics[width=0.30\linewidth]{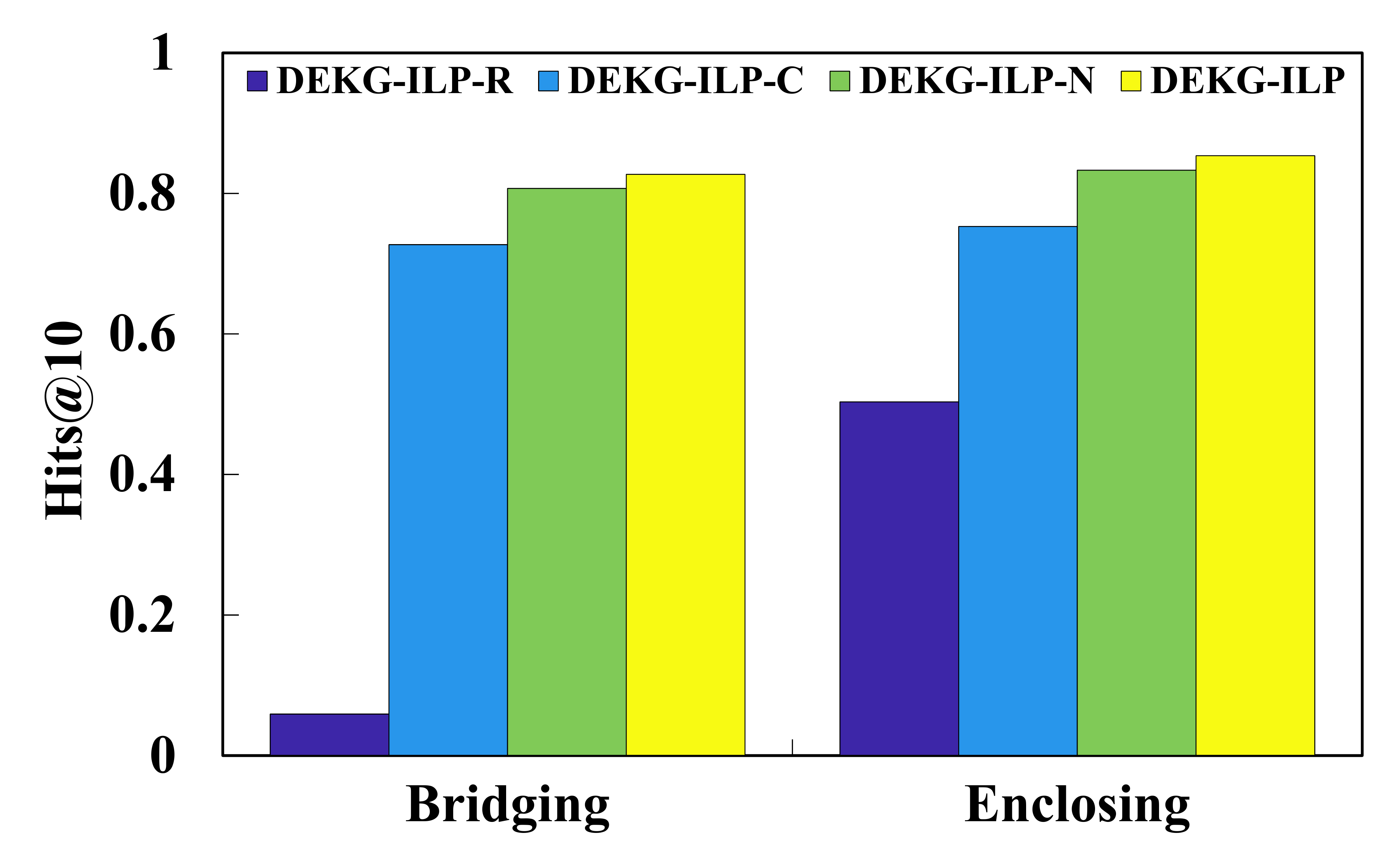}
	}
	\subfigure[FB15k-237 MB]
	{
		\centering
		\includegraphics[width=0.30\linewidth]{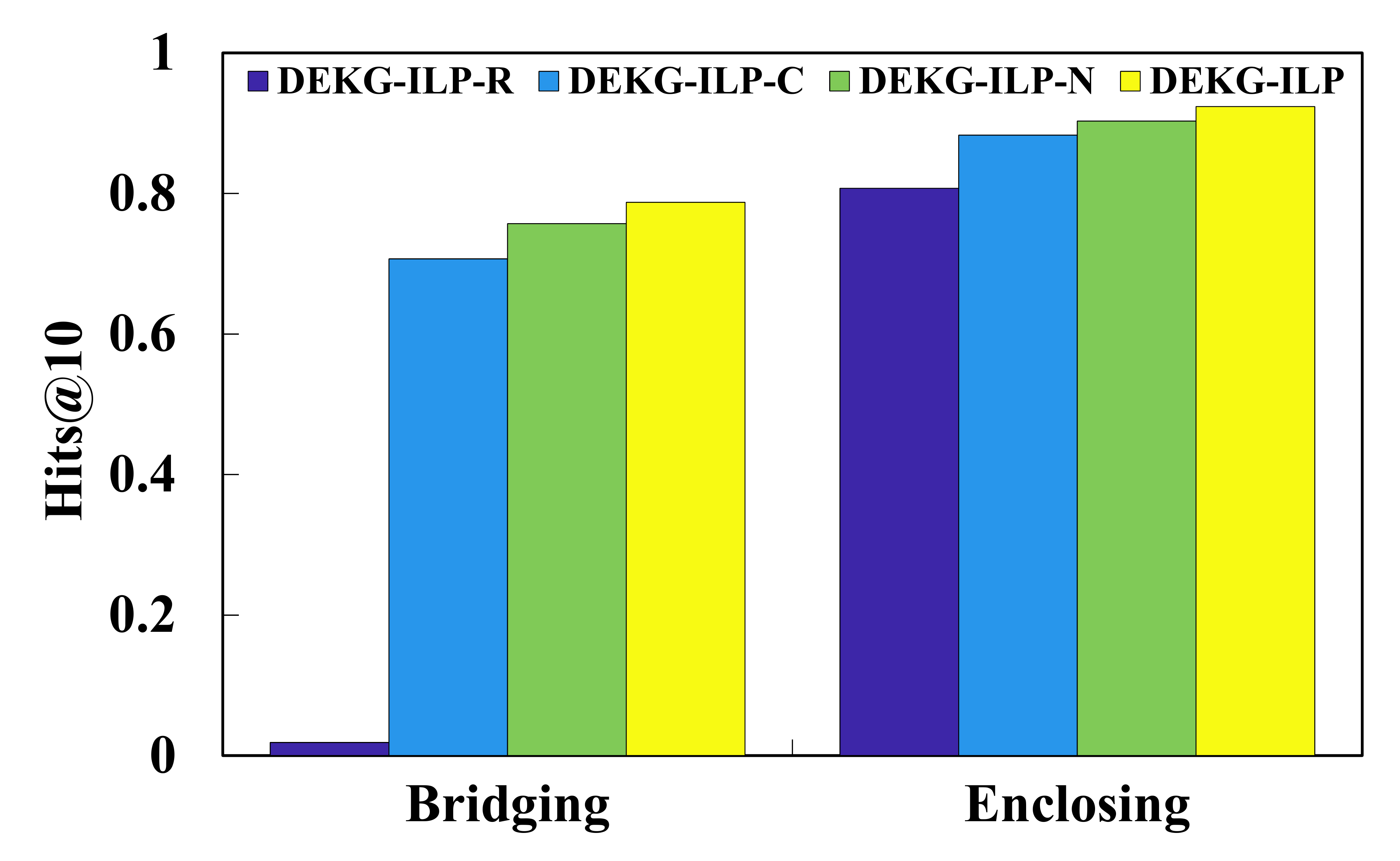}
	}
	\subfigure[FB15k-237 ME]
	{
		\centering
		\includegraphics[width=0.30\linewidth]{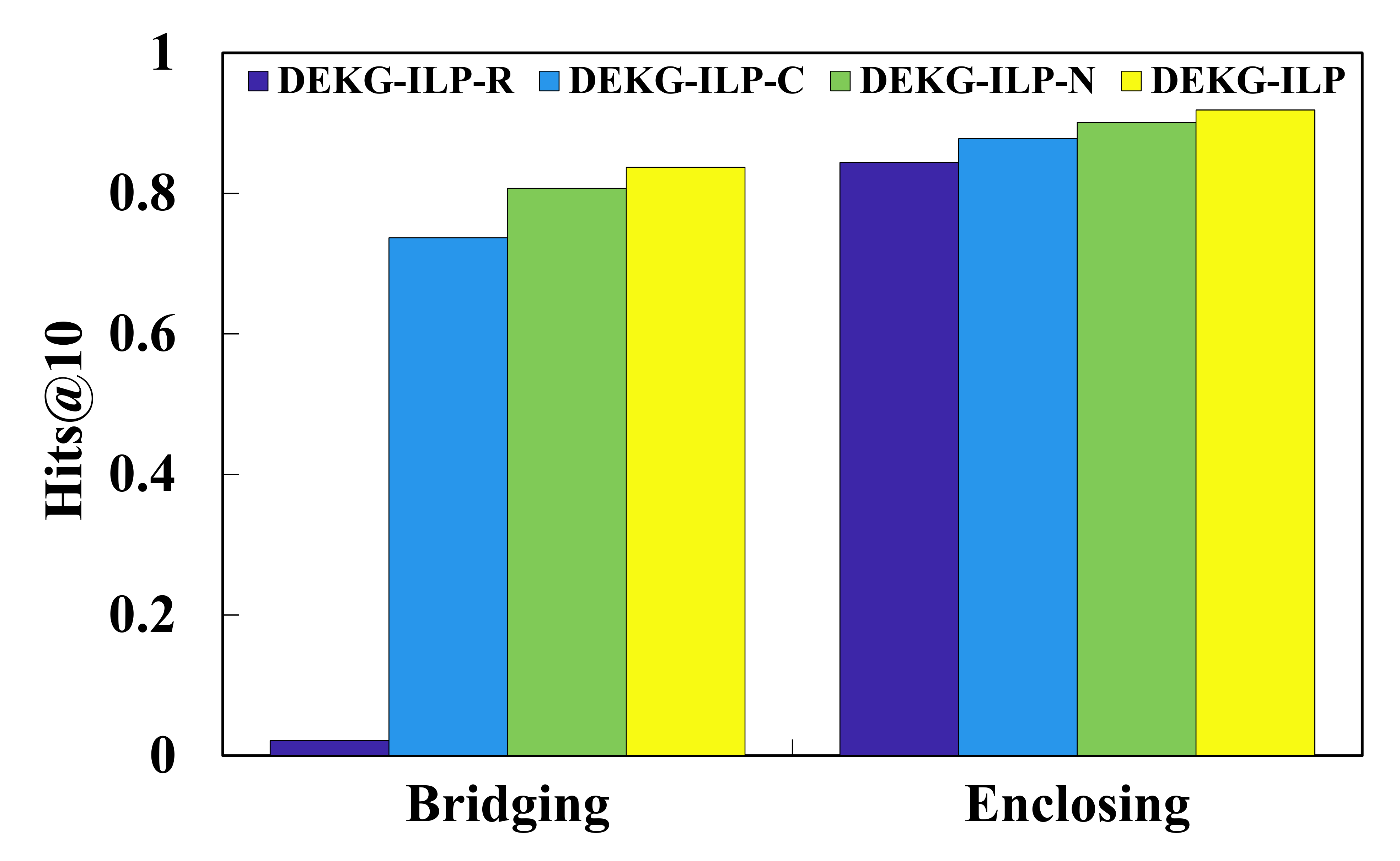}
	} 
	\subfigure[NELL-995 EQ]
	{
		\centering
		\includegraphics[width=0.30\linewidth]{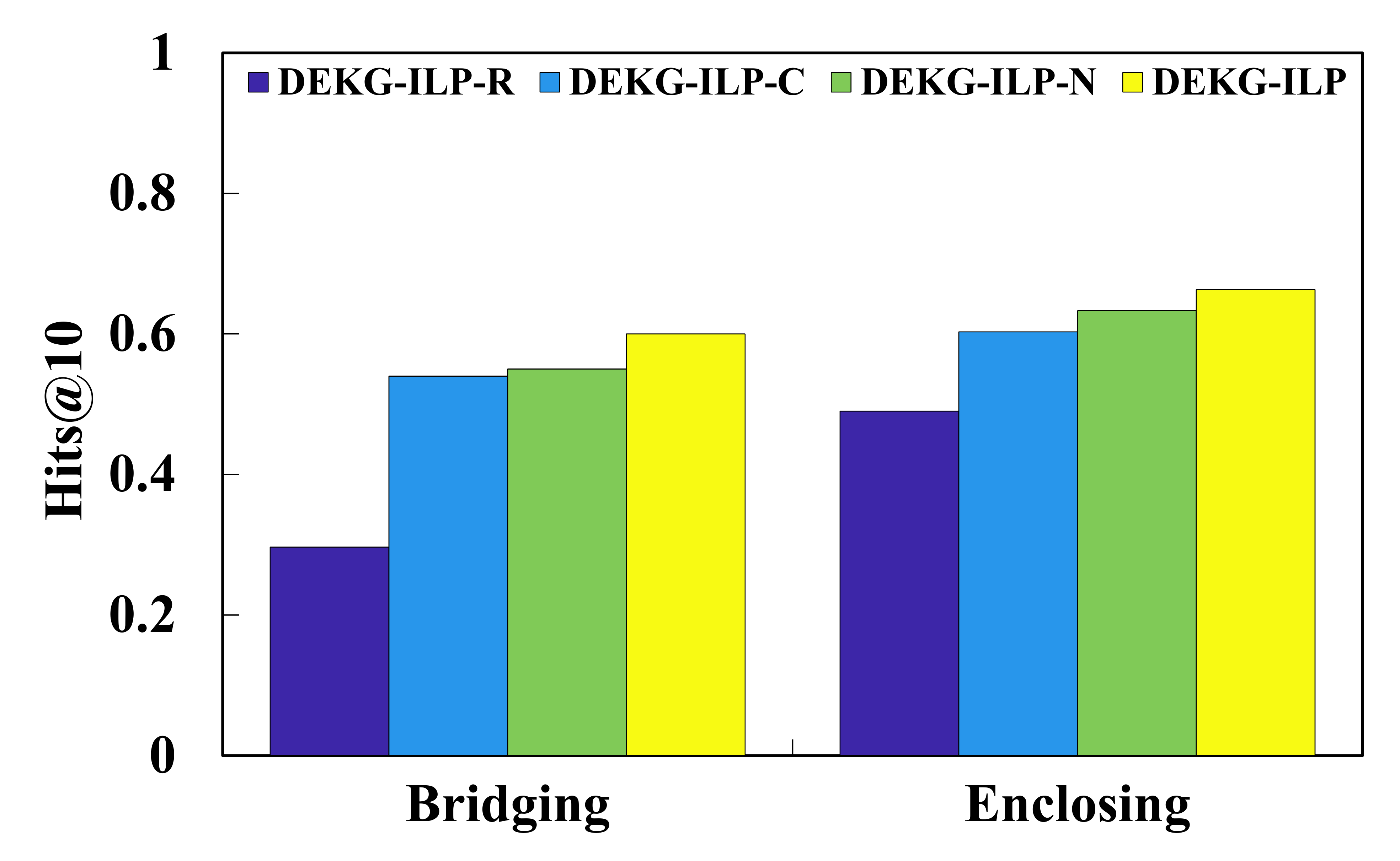}
	}
	\subfigure[NELL-995 MB]
	{
		\centering
		\includegraphics[width=0.30\linewidth]{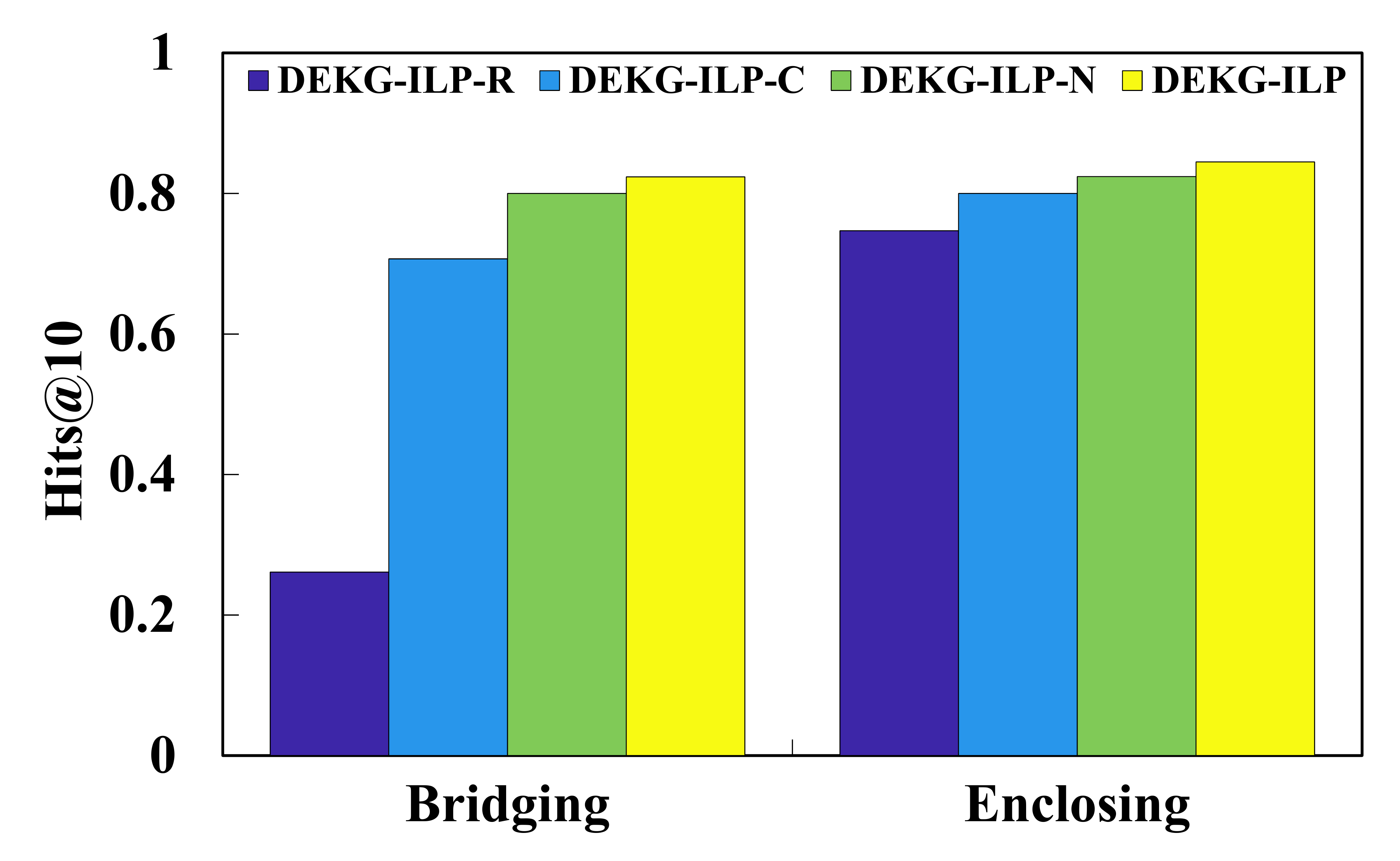}
	} 
	\subfigure[NELL-995 ME]
	{
		\centering
		\includegraphics[width=0.30\linewidth]{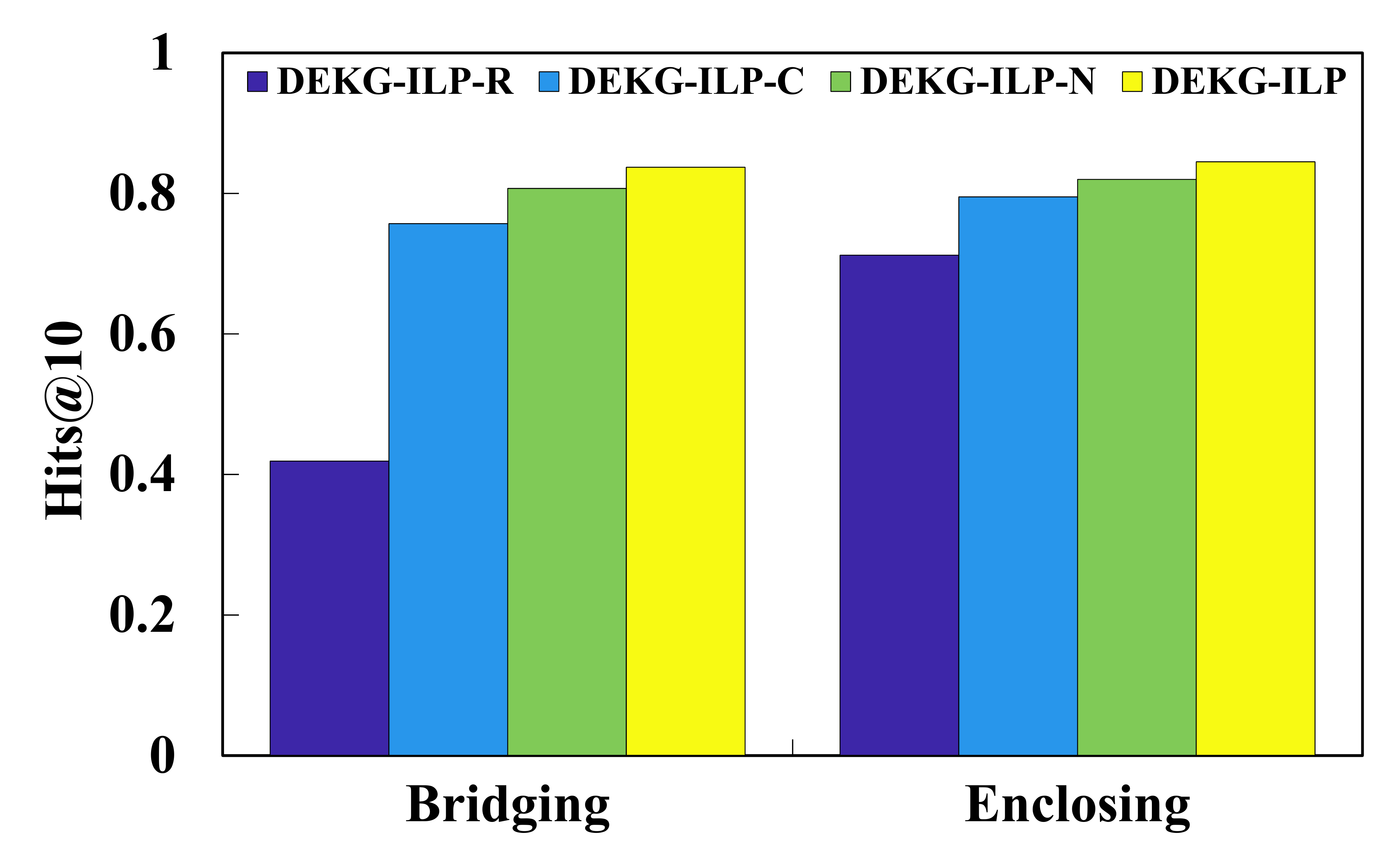}
	}
		\subfigure[WN18RR EQ]
	{
		\centering
		\includegraphics[width=0.30\linewidth]{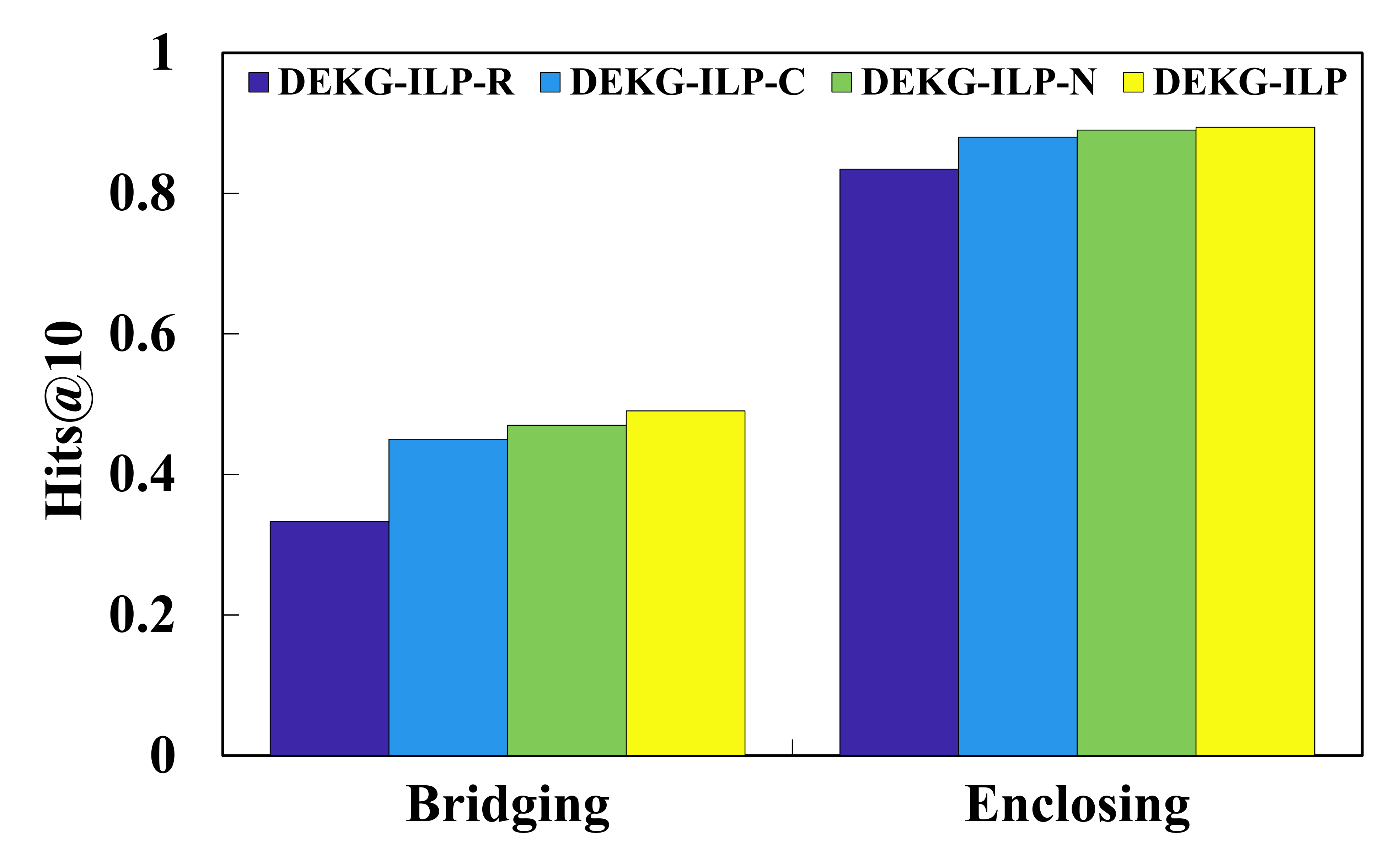}
	}
	\subfigure[WN18RR MB]
	{
		\centering
		\includegraphics[width=0.30\linewidth]{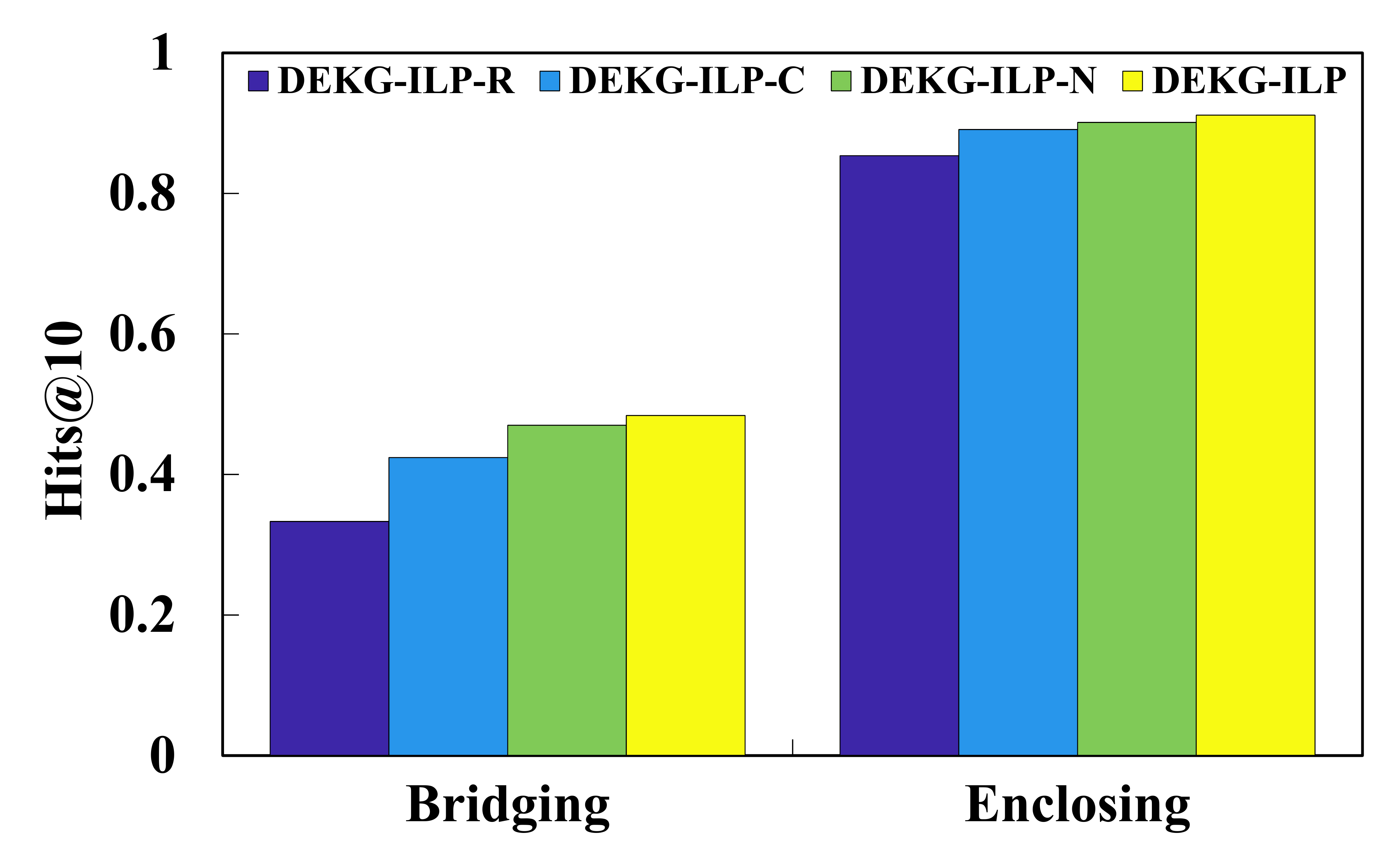}
	} 
	\subfigure[WN18RR ME]
	{
		\centering
		\includegraphics[width=0.30\linewidth]{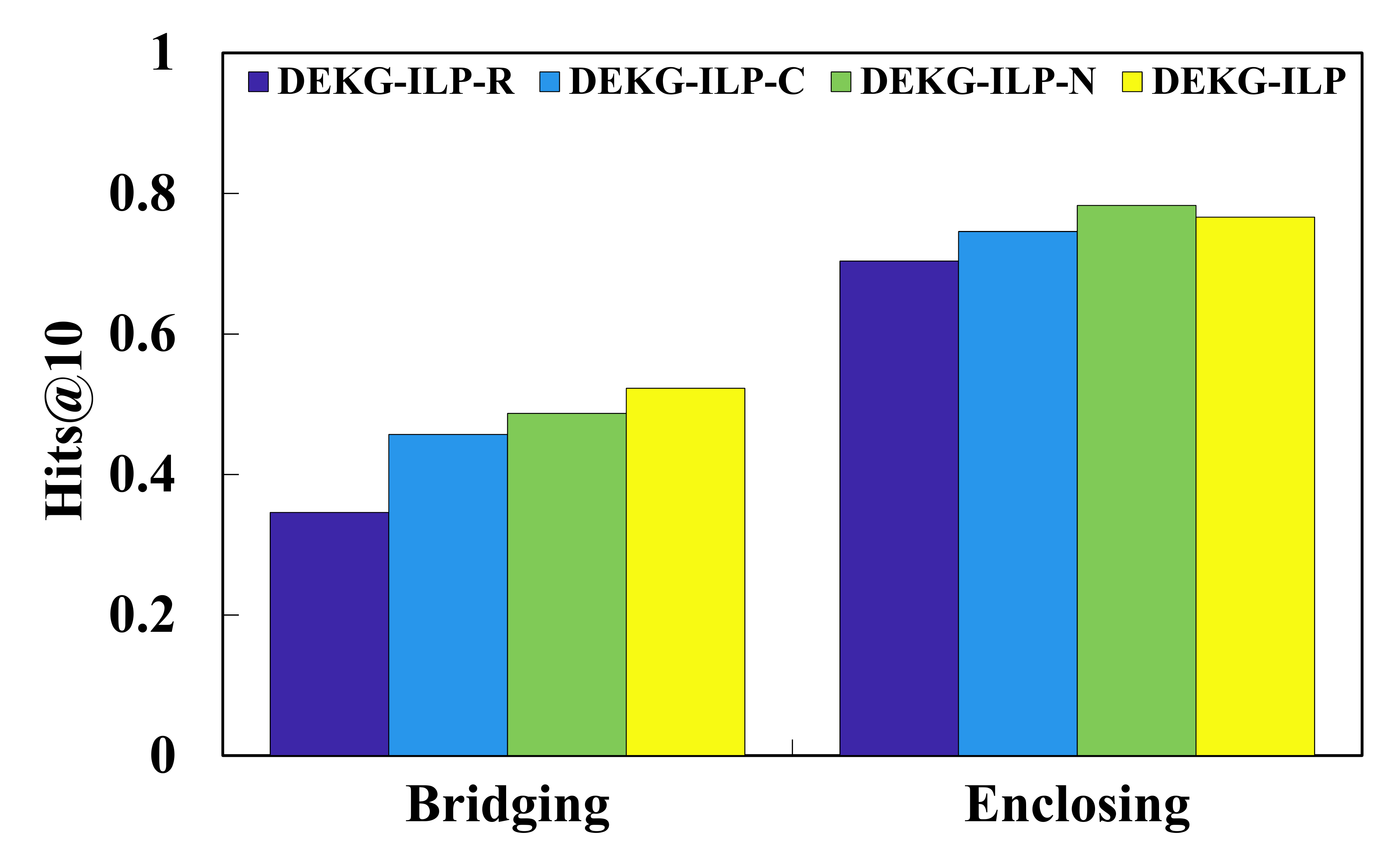}
	}
	\caption{Hits@10 of ablation studies on \emph{EQ}, \emph{MB}, \emph{ME} of FB15k-237, NELL-995, and WN18RR for \emph{enclosing links} and \emph{bridging links} respectively}
	\vspace{-5pt}
	\label{fig:ablation}
\end{figure*}

In this section, we present ablation studies that validate the effectiveness of semantic-aware contrastive learning, relation-specific features in CLRM, and the improved node labeling method in GSM. (1) The variant method DEKG-ILP-R is constructed by removing the semantic score function (i.e., remove $\phi^{sem}$ in Eq.~(\ref{eq:final_score})) to validate the effectiveness of relation-specific features. (2) The variant method DEKG-ILP-C is constructed by removing the contrastive learning loss function (i.e., setting the hyper-parameter $\sigma$ in Eq.~(\ref{eq:learning_object}) to 0) to validate the effectiveness of semantic-aware contrastive learning. (3) The variant method DEKG-ILP-N is constructed by removing the improved node labeling method in GSM to validate the effectiveness of this method.
Fig.~\ref{fig:ablation} presents the Hits@10 results of ablation studies on \emph{EQ}, \emph{MB}, \emph{ME} of three benchmark datasets for \emph{enclosing links} and \emph{bridging links} respectively.

\subsubsection{DEKG-ILP-R} 
The relation-specific features are the key component in CLRM which extract global semantic information in KGs and inherently can be generalized from original KGs to emerging KGs.
The consistent performance gap between DEKG-ILP-R and DEKG-ILP-C when predicting \emph{bridging links} emphasizes the importance of relation-specific features extracted from original KGs. This demonstrates that the global semantic features will not be restricted by the topological structure, i.e., although original KGs and DEKGs are topologically disconnected, the learned relation-based semantic information is still effective for link prediction. 
The improvement can be observed as well when predicting \emph{enclosing links}, demonstrating that the global semantic features also can help to generalize information from original KGs to emerging KGs to predict \emph{enclosing links}. 
What's more, The effectiveness of relation-specific features is more obvious on FB15k-237 and NELL-995 than on WN18RR. It may be caused by the different number of relations that vary in different datasets, demonstrating that DEKG-ILP can extract richer semantic information from the KG if it contain more relations, which align the commonsense.

\subsubsection{DEKG-ILP-C}
The semantic-aware contrastive learning is proposed to optimiaze the relation-specific features in CLRM, with a novel sample strategy to generate positive and negative examples for each entity during training time.
The higher performance of DEKG-ILP compared with DEKG-ILP-C denotes that the proposed semantic-aware contrastive learning method can help to obtain better embeddings of the relation-specific features. This is because the novel sample strategy simulates the semantic variation of entities that we introduced in Section~\ref{sec:clss} and can fully exploit the semantic information in KGs.
It can be observed from the figure that the semantic-aware contrastive learning method has a more obvious improvement for DEKG-ILP on FB15k-237 \emph{MB}, FB15k-237 \emph{ME}, and NELL-995 \emph{ME}. The possible reason is that the entities in the above three datasets have more associated triplets on average (i.e., $|\mathcal{T}| / |\mathcal{E}|$ in TABLE~\ref{tab:datasets}), thus the novel sample strategy proposed in CLRM can generate more diverse positive and negative examples for each entity to optimize the relation-specific features.

\subsubsection{DEKG-ILP-N} 
An improved node labeling method is proposed in GSM to relieve the \emph{topological limitation} in DEKGs. Different from the CLRM that tackles \emph{topological limitation} by extracting features from shared relation space, GSM handles this problem by simulating disconnected nodes using the improved node labeling method. An improvement of around 2\% to 3\% from DEKG-ILP-N to DEKG-ILP can be observed when predicting \emph{bridging links}. However, the improvement is unconspicuous when predicting \emph{enclosing links}, and even backfires on WN18RR \emph{ME}. This is because the subgraph reasoning method in GSM relies on the paths between the head and tail entities. Thus the preserved nodes in the improved node labeling method may become noisy data instead. In general, the improved node labeling method can be helpful when predicting \emph{bridging links}, but is less effective when predicting \emph{enclosing links}.

\begin{figure}
	\centering
	\includegraphics[width=0.85\linewidth]{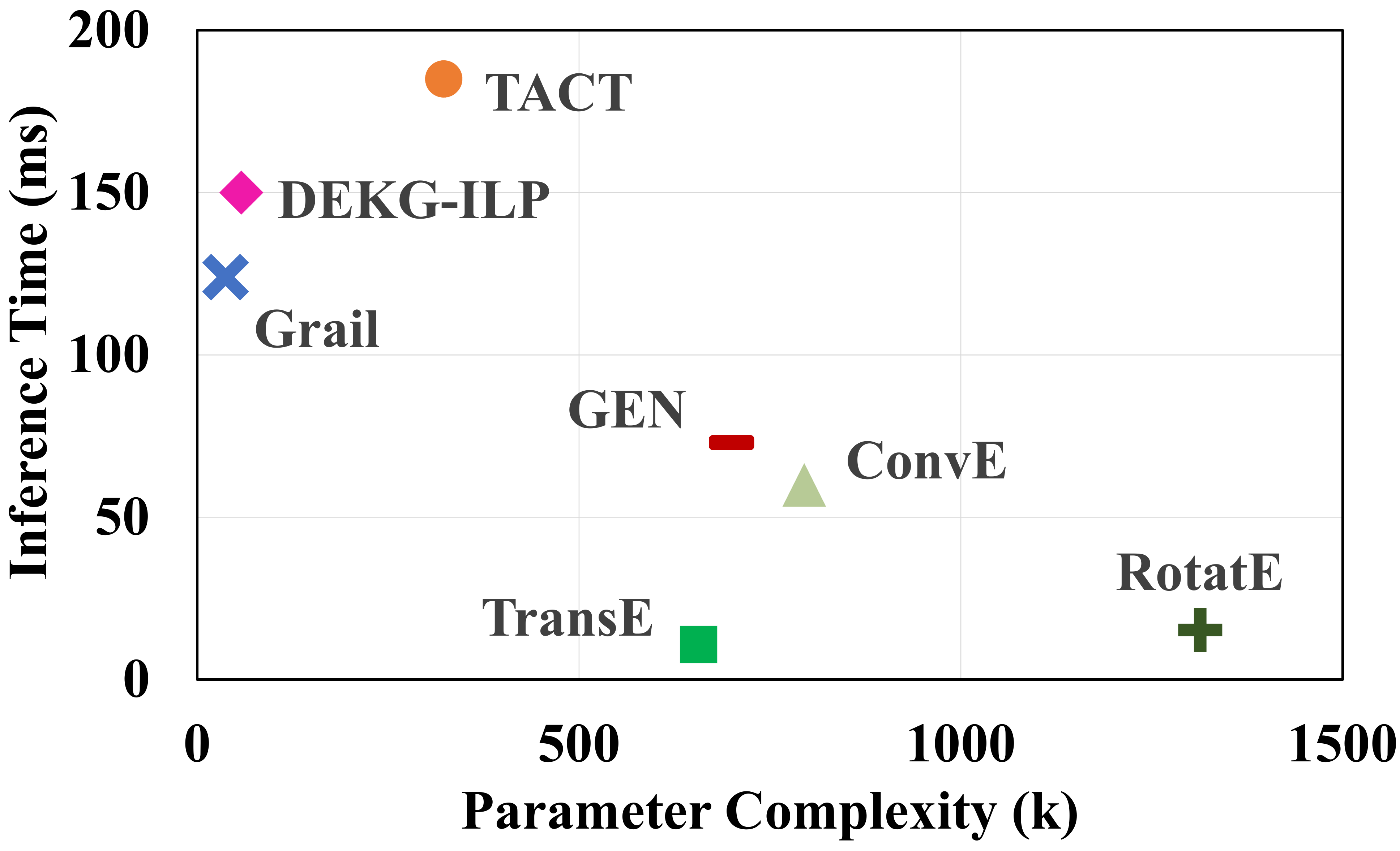}
	\vspace{-5pt}
	\caption{The time and parameter complexity of baselines on FB15k-237 ME.}
	\label{fig:complexity}
	\vspace{-10pt}
\end{figure}

\subsection{Complexity Study}
In this section, we analyze the time and parameter complexity of DEKG-ILP and the compared models. The experiments are conducted on Intel Xeon E5-2650 v4 CPU and single 1080Ti GPU. In Fig.~\ref{fig:complexity}, we report the number of parameters and average inference time for 50 links on FB15k-237 \emph{ME}.

\begin{table*}[t]
\centering
\caption{The training-time and inference-time of each model on all datasets. Epoch denotes the number of training epoch for each model in experiments, T-T denotes training-time (minute) of each epoch, T-I denotes the average inference-time (second) for 50 links.}
\resizebox{\textwidth}{!}
{
    \setlength{\tabcolsep}{1.3mm}{
        \begin{tabular}{cccccccccccccccccccc}
        \toprule
        \multirow{3}{*}{Models} & \multirow{3}{*}{Epoch} & \multicolumn{6}{c}{FB15k-237} & \multicolumn{6}{c}{NELL-995} & \multicolumn{6}{c}{WN18RR} \\
        \cmidrule(r){3-8} \cmidrule(r){9-14} \cmidrule(r){15-20}
                                & & \multicolumn{2}{c}{EQ} & \multicolumn{2}{c}{MB} & \multicolumn{2}{c}{ME} & \multicolumn{2}{c}{EQ} & \multicolumn{2}{c}{MB} & \multicolumn{2}{c}{ME} & \multicolumn{2}{c}{EQ} & \multicolumn{2}{c}{MB} & \multicolumn{2}{c}{ME} \\
        \cmidrule(r){3-4}\cmidrule(r){5-6}\cmidrule(r){7-8}\cmidrule(r){9-10}\cmidrule(r){11-12}\cmidrule(r){13-14}\cmidrule(r){15-16}\cmidrule(r){17-18}\cmidrule(r){19-20}
                                & & T-T & T-I & T-T & T-I & T-T & T-I & T-T & T-I & T-T & T-I & T-T & T-I & T-T & T-I & T-T & T-I & T-T & T-I \\
        \midrule
        TransE                  & 1000  & 0.02 & 0.011 & 0.02 & 0.011 & 0.03 & 0.012 & 0.01 & 0.011 & 0.02 & 0.011 & 0.03 & 0.012 & 0.01 & 0.011 & 0.02 & 0.011 & 0.03 & 0.012 \\
        RotatE                  & 1000  & 0.02 & 0.015 & 0.02 & 0.015 & 0.03 & 0.016 & 0.02 & 0.015 & 0.02 & 0.015 & 0.03 & 0.016 & 0.01 & 0.015 & 0.03 & 0.015 & 0.04 & 0.015 \\
        ConvE                   & 1000  & 0.05 & 0.060 & 0.09 & 0.061 & 0.16 & 0.060 & 0.06 & 0.061 & 0.07 & 0.061 & 0.14 & 0.061 & 0.02 & 0.060 & 0.05 & 0.060 & 0.10 & 0.061 \\
        GEN                     & 5000  & 0.35 & 0.073 & 1.19 & 0.074 & 2.64 & 0.074 & 0.23 & 0.073 & 1.03 & 0.074 & 2.66 & 0.075 & 0.13 & 0.073 & 0.95 & 0.073 & 1.68 & 0.075 \\
        Grail                   & 100   & 4.01 & 0.114 & 13.5 & 0.119 & 31.2 & 0.124 & 1.29 & 0.110 & 7.92 & 0.121 & 24.5 & 0.128 & 0.70 & 0.112 & 1.99 & 0.118 & 3.85 & 0.125 \\
        TACT                    & 100   & 5.72 & 0.170 & 19.4 & 0.177 & 46.3 & 0.185 & 1.85 & 0.165 & 11.4 & 0.180 & 36.7 & 0.186 & 1.40 & 0.172 & 3.46 & 0.176 & 7.62 & 0.181 \\
        DEKG-ILP                & 100   & 4.13 & 0.139 & 14.8 & 0.145 & 32.4 & 0.151 & 1.33 & 0.135 & 8.68 & 0.147 & 25.7 & 0.152 & 0.73 & 0.140 & 2.05 & 0.144 & 3.91 & 0.148 \\
        \bottomrule
        \end{tabular}
    }
}
\vspace{-5pt}
\label{tab:time}
\end{table*}

The parameter complexity of TransE, RotatE, ConvE, and GEN is much higher than Grail, TACT, and DEKG-ILP because these four methods are entity-identify KGE methods where each entity corresponds to an embedding vector, while Grail, TACT, and DEKG-ILP only define learned embeddings for relations in their models. 
The parameter complexity of DEKG-ILP slightly increase compared with Grail but is still much lower than TACT. This is because DEKG-ILP constructs corresponding relation-specific feature for each relation, so its parameter complexity increases to $\mathcal{O}(3|\mathcal{R}|d + 3|\mathcal{R}|dl+2d)$ compared to Grail which is $\mathcal{O}(|\mathcal{R}|d + 3|\mathcal{R}|dl)$, where $|\mathcal{R}|$ is the number of relations, $d$ is the dimension of embedding vector and $l$ is the number of layers in GNNs model. However, TACT models the correlations between relations by considering six different topological interaction of relations thus it's parameter complexity grows to $\mathcal{O}(7|\mathcal{R}|d +3|\mathcal{R}|dl + |\mathcal{R}|^2+2d^2)$. 

Although with a smaller parameter size, the time complexity of the subgraph reasoning methods (i.e., Grail, TACT, and DEKG-ILP) are generally higher than that of entity-identify KGE methods (i.e., TransE, ConvE, RotatE, GEN). This is because the subgraph reasoning methods involve a more complex GNNs-based encoder and a subgraph extracting process using the shortest path algorithm, whose time complexity is $\mathcal{O}(log(\mathcal{V})\mathcal{E} + \mathcal{R}dk$) \cite{Grail}, while TransE and RotatE directly calculate the score for each link with the entity embeddings, thus the corresponding time complexity decrease to $\mathcal{O}(d)$. The time complexity of ConvE and GEN is a litte higher than that of TransE and RotatE, since they introduce a CNN and a GNN as the encoder respectively. The details of training-time and inference-time are presented in TABLE~\ref{tab:time}.

In summary, our proposed model DEKG-ILP has a significant advantage in model parameter complexity although it sacrifices some efficiency, but the overall running time is still tolerable in practice (i.e., 145ms for 50 links on average).
\begin{figure}
	\centering
	\subfigbottomskip=2pt
	\subfigure[The heat map of the enclosing link in FB15k-237]
	{
		\centering
		\includegraphics[width=0.9\linewidth]{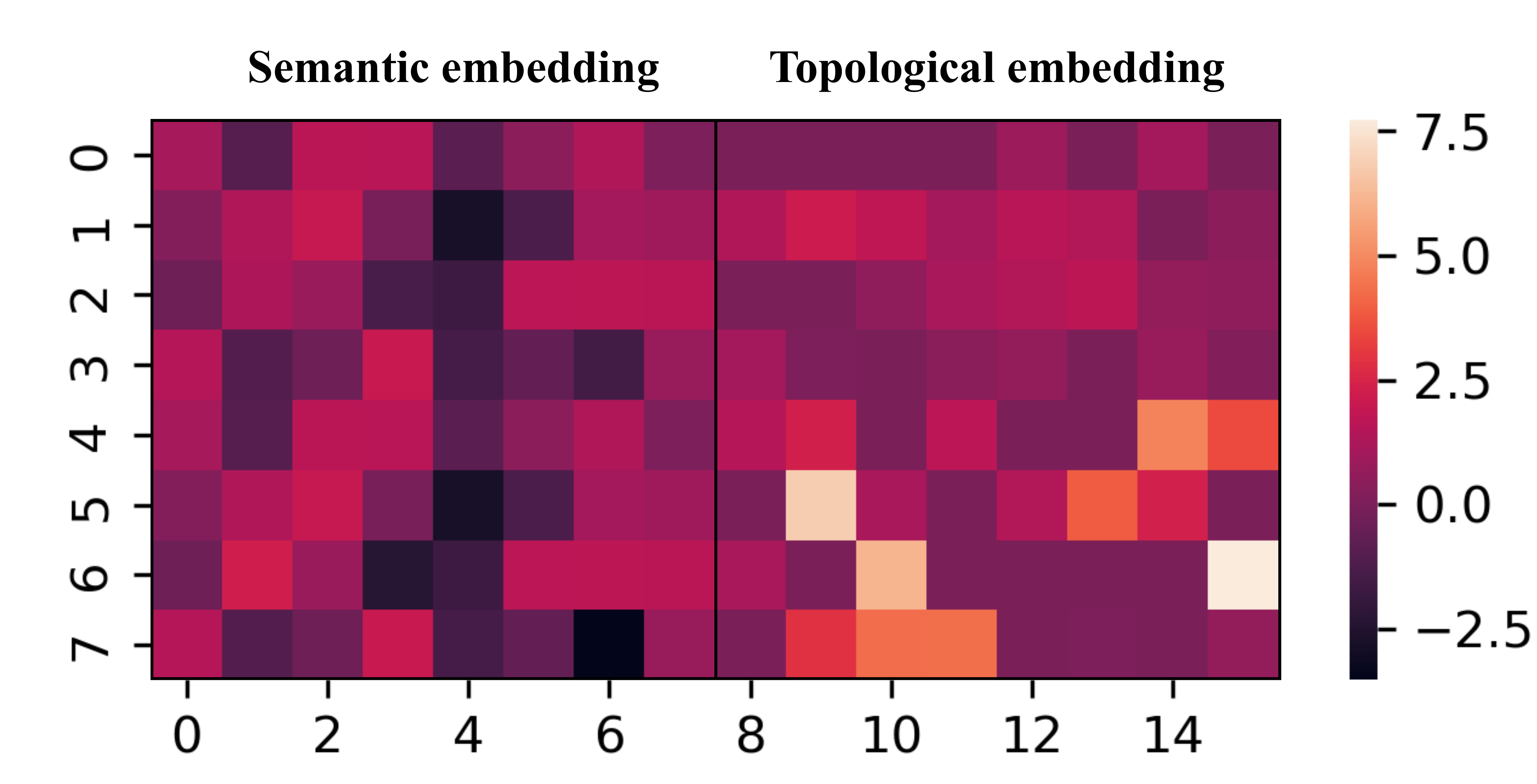}
	}
	\subfigure[The heat map of the bridging link in NELL-995]
	{
		\centering
		\includegraphics[width=0.9\linewidth]{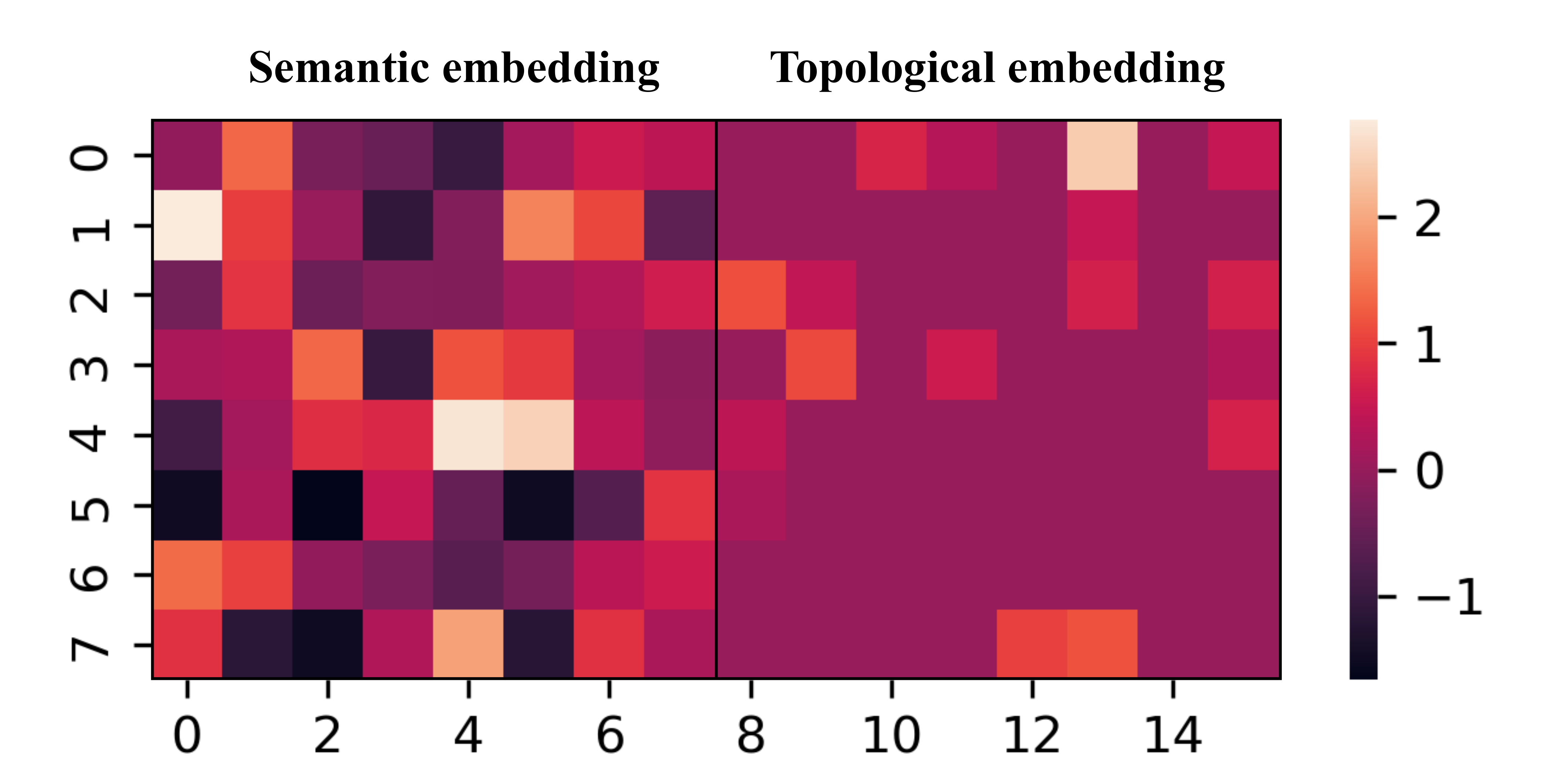}
	}
	\caption{The embedding heap maps of the enclosing link and bridging link}
	\label{fig:heatmap}
	\vspace{-10pt}
\end{figure}

\subsection{Case Study}
In this section, we choose an \emph{enclosing link} \emph{(08720, film production\_companies, 0g1rw)} in FB15k-237 and a \emph{bridging link} \emph{(spurs, team\_play\_against\_team, grizzlies)} in NELL-995 to visualize how the encoder in DEKG-ILP works on \emph{enclosing links} and \emph{bridging links} respectively.
The embedding heat maps of the chosen links are shown in Fig.~\ref{fig:heatmap}. To construct the heat maps, we firstly concatenate and resize the 32-dimensional embeddings $\bold{e}_i$ and $\bold{e}_j$ in $\phi^{sem}$ in CLRM as an $8 \times 8$ matrix. Then we visualize this matrix as the semantic embedding heat map in the left of Fig.~\ref{fig:heatmap}. The same operation is performed on $\boldsymbol{h}_i^L$ and $\boldsymbol{h}_j^L$ in $\phi^{tpo}$ in GSM to obtain the topological embedding heat map in the right. Observed from Fig.~\ref{fig:heatmap}(b), when predicting the \emph{bridging link}, there are many active values in the semantic embedding, while most values in topological embedding are close to zero. However, as shown in Fig.~\ref{fig:heatmap}(a), the distribution of active values is much more balanced when predicting the \emph{enclosing link}. This demonstrates that the module CLRM plays a more important role than GSM when predicting \emph{bridging links}, but the contribution of the two modules is similar when predicting \emph{enclosing links}, enabling DEKG-ILP to have a good performance on predicting both enclosing and bridging links.

\section{Conclusion}
\label{sec:conclusion}
In this paper, we extend the problem of inductive link prediction for a disconnected emerging knowledge graph to consider \emph{enclosing links} and especially \emph{bridging links} in DEKGs.
To handle this problem, we propose a novel model entitled DEKG-ILP which contains two modules CLRM and GSM.  Specifically, the Contrastive Learning-based Relation-specific Feature Modeling module CLRM is employed to exploit the global semantic features shared by original KGs and DEKGs, where a semantic-aware contrastive learning method with a novel sampling strategy is designed to optimize these features. The GNN-based Subgraph Modeling module GSM with an improved node labeling method is used to exploit the local topological information in KGs. 
Comprehensive experiments have been done. The results demonstrate that our proposed model not only have a better performance on predicting \emph{enclosing links}, but also can handle the problem of \emph{bridging links} that ignored by previous work.

\section*{Acknowledgment}
This work was supported by the National Natural Science Foundation of China (No. 61902270), 
the Major Program of the Natural Science Foundation of Jiangsu Higher Education Institutions of China (No. 19KJA610002),
Australian Research Council (Nos.FT210100624, DP190101985).

\bibliographystyle{IEEEtranS} 
\balance
\bibliography{IEEEabrv}

\begin{thebibliography}{10}
\providecommand{\url}[1]{#1}
\csname url@samestyle\endcsname
\providecommand{\newblock}{\relax}
\providecommand{\bibinfo}[2]{#2}
\providecommand{\BIBentrySTDinterwordspacing}{\spaceskip=0pt\relax}
\providecommand{\BIBentryALTinterwordstretchfactor}{4}
\providecommand{\BIBentryALTinterwordspacing}{\spaceskip=\fontdimen2\font plus
\BIBentryALTinterwordstretchfactor\fontdimen3\font minus
  \fontdimen4\font\relax}
\providecommand{\BIBforeignlanguage}[2]{{%
\expandafter\ifx\csname l@#1\endcsname\relax
\typeout{** WARNING: IEEEtranS.bst: No hyphenation pattern has been}%
\typeout{** loaded for the language `#1'. Using the pattern for}%
\typeout{** the default language instead.}%
\else
\language=\csname l@#1\endcsname
\fi
#2}}
\providecommand{\BIBdecl}{\relax}
\BIBdecl

\bibitem{DBpedia}
S.~Auer, C.~Bizer, G.~Kobilarov, J.~Lehmann, R.~Cyganiak, and Z.~G. Ives,
  ``Dbpedia: {A} nucleus for a web of open data,'' in \emph{The Semantic Web},
  2007, pp. 722--735.

\bibitem{GEN}
J.~Baek, D.~B. Lee, and S.~J. Hwang, ``Learning to extrapolate knowledge:
  Transductive few-shot out-of-graph link prediction,'' in \emph{NeurIPS},
  2020, pp. 6--12.

\bibitem{Freebase}
K.~D. Bollacker, C.~Evans, P.~Paritosh, T.~Sturge, and J.~Taylor, ``Freebase: a
  collaboratively created graph database for structuring human knowledge,'' in
  \emph{SIGMOD}, 2008, pp. 1247--1250.

\bibitem{TransE}
A.~Bordes, N.~Usunier, A.~Garc{\'{\i}}a{-}Dur{\'{a}}n, J.~Weston, and
  O.~Yakhnenko, ``Translating embeddings for modeling multi-relational data,''
  in \emph{NIPS}, 2013, pp. 2787--2795.

\bibitem{NELL}
A.~Carlson, J.~Betteridge, B.~Kisiel, B.~Settles, E.~R.~H. Jr., and T.~M.
  Mitchell, ``Toward an architecture for never-ending language learning,'' in
  \emph{AAAI}, 2010, pp. 1306--1313.

\bibitem{TACT}
J.~Chen, H.~He, F.~Wu, and J.~Wang, ``Topology-aware correlations between
  relations for inductive link prediction in knowledge graphs,'' in
  \emph{AAAI}, 2021, pp. 6271--6278.

\bibitem{Cuco}
G.~Chu, X.~Wang, C.~Shi, and X.~Jiang, ``Cuco: Graph representation with
  curriculum contrastive learning,'' in \emph{IJCAI}, 2021, pp. 2300--2306.

\bibitem{TensorLog}
W.~W. Cohen, ``Tensorlog: {A} differentiable deductive database,'' \emph{CoRR},
  vol. abs/1605.06523, 2016.

\bibitem{ConvE}
T.~Dettmers, P.~Minervini, P.~Stenetorp, and S.~Riedel, ``Convolutional 2d
  knowledge graph embeddings,'' in \emph{AAAI}, 2018, pp. 1811--1818.

\bibitem{AMIE}
L.~A. Gal{\'{a}}rraga, C.~Teflioudi, K.~Hose, and F.~M. Suchanek, ``{AMIE:}
  association rule mining under incomplete evidence in ontological knowledge
  bases,'' in \emph{WWW}, 2013, pp. 413--422.

\bibitem{MEAN}
T.~Hamaguchi, H.~Oiwa, M.~Shimbo, and Y.~Matsumoto, ``Knowledge transfer for
  out-of-knowledge-base entities : {A} graph neural network approach,'' in
  \emph{IJCAI}, 2017, pp. 1802--1808.

\bibitem{openke}
X.~Han, S.~Cao, L.~Xin, Y.~Lin, Z.~Liu, M.~Sun, and J.~Li, ``Openke: An open
  toolkit for knowledge embedding,'' in \emph{EMNLP}, 2018.

\bibitem{MoCo}
K.~He, H.~Fan, Y.~Wu, S.~Xie, and R.~B. Girshick, ``Momentum contrast for
  unsupervised visual representation learning,'' in \emph{CVPR}, 2020, pp.
  9726--9735.

\bibitem{VNnetwork}
Y.~He, Z.~Wang, P.~Zhang, Z.~Tu, and Z.~Ren, ``{VN} network: Embedding newly
  emerging entities with virtual neighbors,'' in \emph{CIKM}, 2020, pp.
  505--514.

\bibitem{DeepInfomax}
R.~D. Hjelm, A.~Fedorov, S.~Lavoie{-}Marchildon, K.~Grewal, P.~Bachman,
  A.~Trischler, and Y.~Bengio, ``Learning deep representations by mutual
  information estimation and maximization,'' in \emph{ICLR}, 2019.

\bibitem{KG_QA_1}
S.~Hu, L.~Zou, J.~X. Yu, H.~Wang, and D.~Zhao, ``Answering natural language
  questions by subgraph matching over knowledge graphs (extended abstract),''
  in \emph{ICDE}, 2018, pp. 1815--1816.

\bibitem{add1}
N.~Q.~V. Hung, C.~T. Duong, T.~T. Nguyen, M.~Weidlich, K.~Aberer, H.~Yin, and
  X.~Zhou, ``Argument discovery via crowdsourcing,'' \emph{{VLDB} J.}, vol.~26,
  no.~4, pp. 511--535, 2017.

\bibitem{KG_QA_2}
M.~Kaiser, R.~S. Roy, and G.~Weikum, ``Reinforcement learning from
  reformulations in conversational question answering over knowledge graphs,''
  in \emph{SIGIR}, 2021, pp. 459--469.

\bibitem{TransN}
Z.~Li, W.~Zheng, X.~Lin, Z.~Zhao, Z.~Wang, Y.~Wang, X.~Jian, L.~Chen, Q.~Yan,
  and T.~Mao, ``Transn: Heterogeneous network representation learning by
  translating node embeddings,'' in \emph{ICDE}, 2020, pp. 589--600.

\bibitem{RuleN}
C.~Meilicke, M.~Fink, Y.~Wang, D.~Ruffinelli, R.~Gemulla, and
  H.~Stuckenschmidt, ``Fine-grained evaluation of rule- and embedding-based
  systems for knowledge graph completion,'' in \emph{ISWC}, 2018, pp. 3--20.

\bibitem{word2vector}
T.~Mikolov, I.~Sutskever, K.~Chen, G.~S. Corrado, and J.~Dean, ``Distributed
  representations of words and phrases and their compositionality,'' in
  \emph{NIPS}, 2013, pp. 3111--3119.

\bibitem{traditional_rule}
S.~Muggleton, ``Inductive logic programming,'' \emph{New Gener. Comput.},
  vol.~8, no.~4, pp. 295--318, 1991.

\bibitem{ConvKB}
D.~Q. Nguyen, T.~D. Nguyen, D.~Q. Nguyen, and D.~Q. Phung, ``A novel embedding
  model for knowledge base completion based on convolutional neural network,''
  in \emph{NAACL-HLT}, 2018, pp. 327--333.

\bibitem{add3}
T.~T. Nguyen, M.~Weidlich, D.~C. Thang, H.~Yin, and N.~Q.~V. Hung, ``Retaining
  data from streams of social platforms with minimal regret,'' in \emph{IJCAI},
  C.~Sierra, Ed., 2017, pp. 2850--2856.

\bibitem{RESCAL}
M.~Nickel, V.~Tresp, and H.~Kriegel, ``A three-way model for collective
  learning on multi-relational data,'' in \emph{ICML}, 2011, pp. 809--816.

\bibitem{RLvRL}
P.~G. Omran, K.~Wang, and Z.~Wang, ``An embedding-based approach to rule
  learning in knowledge graphs,'' \emph{TKDE}, vol.~33, no.~4, pp. 1348--1359,
  2021.

\bibitem{GCC}
J.~Qiu, Q.~Chen, Y.~Dong, J.~Zhang, H.~Yang, M.~Ding, K.~Wang, and J.~Tang,
  ``{GCC:} graph contrastive coding for graph neural network pre-training,'' in
  \emph{SIGKDD}, 2020, pp. 1150--1160.

\bibitem{DURM}
A.~Sadeghian, M.~Armandpour, P.~Ding, and D.~Z. Wang, ``Drum: End-to-end
  differentiable rule mining on knowledge graphs.'' in \emph{NeurIPS}, 2019,
  pp. 15\,321--15\,331.

\bibitem{R-GCN}
M.~S. Schlichtkrull, T.~N. Kipf, P.~Bloem, R.~van~den Berg, I.~Titov, and
  M.~Welling, ``Modeling relational data with graph convolutional networks,''
  in \emph{ESWC}, 2018, pp. 593--607.

\bibitem{SACN}
C.~Shang, Y.~Tang, J.~Huang, J.~Bi, X.~He, and B.~Zhou, ``End-to-end
  structure-aware convolutional networks for knowledge base completion,'' in
  \emph{AAAI}, 2019, pp. 3060--3067.

\bibitem{ConMask}
B.~Shi and T.~Weninger, ``Open-world knowledge graph completion,'' in
  \emph{AAAI}, 2018, pp. 1957--1964.

\bibitem{MoCL}
M.~Sun, J.~Xing, H.~Wang, B.~Chen, and J.~Zhou, ``Mocl: Data-driven molecular
  fingerprint via knowledge-aware contrastive learning from molecular graph,''
  in \emph{SIGKDD}, 2021, pp. 3585--3594.

\bibitem{RotatE}
Z.~Sun, Z.~Deng, J.~Nie, and J.~Tang, ``Rotate: Knowledge graph embedding by
  relational rotation in complex space,'' in \emph{ICLR}, 2019.

\bibitem{Grail}
K.~K. Teru, E.~Denis, and W.~Hamilton, ``Inductive relation prediction by
  subgraph reasoning,'' in \emph{ICML}, 2020, pp. 9448--9457.

\bibitem{WN18RR}
K.~Toutanova and D.~Chen, ``Observed versus latent features for knowledge base
  and text inference,'' in \emph{CVSC}, 2015, pp. 57--66.

\bibitem{ComplEx}
T.~Trouillon, J.~Welbl, S.~Riedel, {\'{E}}.~Gaussier, and G.~Bouchard,
  ``Complex embeddings for simple link prediction,'' in \emph{ICML}, 2016, pp.
  2071--2080.

\bibitem{DGI}
P.~Velickovic, W.~Fedus, W.~L. Hamilton, P.~Li{\`{o}}, Y.~Bengio, and R.~D.
  Hjelm, ``Deep graph infomax,'' in \emph{ICLR}, 2019.

\bibitem{LAN}
P.~Wang, J.~Han, C.~Li, and R.~Pan, ``Logic attention based neighborhood
  aggregation for inductive knowledge graph embedding,'' in \emph{AAAI}, 2019,
  pp. 7152--7159.

\bibitem{add5}
Q.~Wang, H.~Yin, W.~Wang, Z.~Huang, G.~Guo, and Q.~V.~H. Nguyen, ``Multi-hop
  path queries over knowledge graphs with neural memory networks,'' in
  \emph{DASFAA}, ser. Lecture Notes in Computer Science, vol. 11446, 2019, pp.
  777--794.

\bibitem{KG_review}
Q.~Wang, Z.~Mao, B.~Wang, and L.~Guo, ``Knowledge graph embedding: {A} survey
  of approaches and applications,'' \emph{TKDE}, vol.~29, no.~12, pp.
  2724--2743, 2017.

\bibitem{KG_IR_1}
Y.~Wang, A.~Khan, T.~Wu, J.~Jin, and H.~Yan, ``Semantic guided and response
  times bounded top-k similarity search over knowledge graphs,'' in
  \emph{ICDE}, 2020, pp. 445--456.

\bibitem{TransH}
Z.~Wang, J.~Zhang, J.~Feng, and Z.~Chen, ``Knowledge graph embedding by
  translating on hyperplanes,'' in \emph{AAAI}, 2014, pp. 1112--1119.

\bibitem{IKRL}
R.~Xie, Z.~Liu, H.~Luan, and M.~Sun, ``Image-embodied knowledge representation
  learning,'' in \emph{IJCAI}, 2017, pp. 3140--3146.

\bibitem{TKRL}
R.~Xie, Z.~Liu, and M.~Sun, ``Representation learning of knowledge graphs with
  hierarchical types,'' in \emph{IJCAI}, 2016, pp. 2965--2971.

\bibitem{NELL-995}
W.~Xiong, T.~Hoang, and W.~Y. Wang, ``Deeppath: {A} reinforcement learning
  method for knowledge graph reasoning,'' in \emph{EMNLP}, 2017, pp. 564--573.

\bibitem{DistMult}
B.~Yang, W.~Yih, X.~He, J.~Gao, and L.~Deng, ``Embedding entities and relations
  for learning and inference in knowledge bases,'' in \emph{ICLR}, 2015, pp.
  1--12.

\bibitem{NeuralLP}
F.~Yang, Z.~Yang, and W.~W. Cohen, ``Differentiable learning of logical rules
  for knowledge base reasoning,'' in \emph{NIPS}, 2017, pp. 2319--2328.

\bibitem{KG_IR_2}
Z.~Yang, ``Biomedical information retrieval incorporating knowledge graph for
  explainable precision medicine,'' in \emph{SIGIR}, J.~Huang, Y.~Chang,
  X.~Cheng, J.~Kamps, V.~Murdock, J.~Wen, and Y.~Liu, Eds., 2020, p. 2486.

\bibitem{add4}
J.~Yu, H.~Yin, J.~Li, M.~Gao, Z.~Huang, and L.~Cui, ``Enhance social
  recommendation with adversarial graph convolutional networks,'' \emph{TKDE},
  2020.

\bibitem{DBLP:conf/aaai/ZengX21}
J.~Zeng and P.~Xie, ``Contrastive self-supervised learning for graph
  classification,'' in \emph{AAAI}, 2021, pp. 10\,824--10\,832.

\bibitem{HAKE}
Z.~Zhang, J.~Cai, Y.~Zhang, and J.~Wang, ``Learning hierarchy-aware knowledge
  graph embeddings for link prediction,'' in \emph{AAAI}, 2020, pp. 3065--3072.

\bibitem{add2}
K.~Zhao, Y.~Zhang, H.~Yin, J.~Wang, K.~Zheng, X.~Zhou, and C.~Xing,
  ``Discovering subsequence patterns for next {POI} recommendation,'' in
  \emph{IJCAI}, 2020, pp. 3216--3222.

\bibitem{contrastive_NLP1}
M.~Zhou, Z.~Li, and P.~Xie, ``Self-supervised regularization for text
  classification,'' \emph{Trans. Assoc. Comput. Linguistics}, vol.~9, pp.
  641--656, 2021.

\end{thebibliography}

\end{document}